\documentclass{article}

\usepackage{microtype}
\usepackage{graphicx}
\usepackage{subfigure}
\usepackage{booktabs} 
\usepackage{multirow}
\usepackage{hyperref}
\usepackage{caption}
\usepackage{subcaption}
\usepackage{subfigure}

\usepackage[accepted]{icml2025}

\usepackage{amsmath}
\usepackage{amssymb}
\usepackage{mathtools}
\usepackage{amsthm}

\usepackage[capitalize,noabbrev]{cleveref}

\theoremstyle{plain}

\theoremstyle{definition}

\theoremstyle{remark}

\usepackage[textsize=tiny]{todonotes}

\icmltitlerunning{SpectralKD: A Unified Framework for Interpreting and Distilling Vision Transformers via Spectral Analysis}

\begin{document}

\twocolumn[
\icmltitle{SpectralKD: A Unified Framework for Interpreting and \\
	Distilling Vision Transformers via Spectral Analysis}



\icmlsetsymbol{equal}{*}

\begin{icmlauthorlist}
\icmlauthor{Huiyuan Tian}{yyy}
\icmlauthor{Bonan Xu}{sch}
\icmlauthor{Shijian Li}{yyy}
\icmlauthor{Gang Pan}{yyy}
\end{icmlauthorlist}

\icmlaffiliation{yyy}{College of Computer Science and Technology, Zhejiang University, NO. 38 Zheda Road, Xihu District, Hangzhou 310027, China}
\icmlaffiliation{sch}{School of Aeronautics and Astronautics, Zhejiang University, NO. 38 Zheda Road, Xihu District, Hangzhou 310027, China}

\icmlcorrespondingauthor{Shijian Li}{shijianli@zju.edu.cn}

\icmlkeywords{Machine Learning, ICML}

\vskip 0.3in
]



\printAffiliationsAndNotice{}  

\begin{abstract}
	Knowledge Distillation (KD) has achieved widespread success in compressing large Vision Transformers (ViTs), but a unified theoretical framework for both ViTs and KD is still lacking. In this paper, we propose SpectralKD, a novel unified analytical framework that offers deeper insights into ViTs and optimizes KD via spectral analysis. Our model-wise analysis reveals that CaiT concentrates information in their first and last few layers, informing optimal layer selection for KD. Surprisingly, our layer-wise analysis discovers that Swin Transformer and CaiT exhibit similar spectral encoding patterns despite their architectural differences, leading to feature map alignment guideline. Building on these insights, we propose a simple yet effective spectral alignment method for KD. Benefiting from the deeper understanding by above analysis results, even such a simple strategy achieves state-of-the-art performance on ImageNet-1K without introducing any trainable parameters, improving DeiT-Tiny by $+5.2\%$ and Swin-Tiny by $+1.4\%$ in top-1 accuracy. Furthermore, our post-training analysis reveals that distilled students can reproduce spectral patterns similar to their teachers, opening a new area we term ``distillation dynamics". Code and experimental logs are available in \url{ https://github.com/thy960112/SpectralKD}.
\end{abstract}

\begin{figure}[t]
	\centering
	\begin{minipage}[b]{\linewidth}
		\subfigure[CaiT-S24 (teacher).]{
			
			\includegraphics[width=0.98\linewidth]{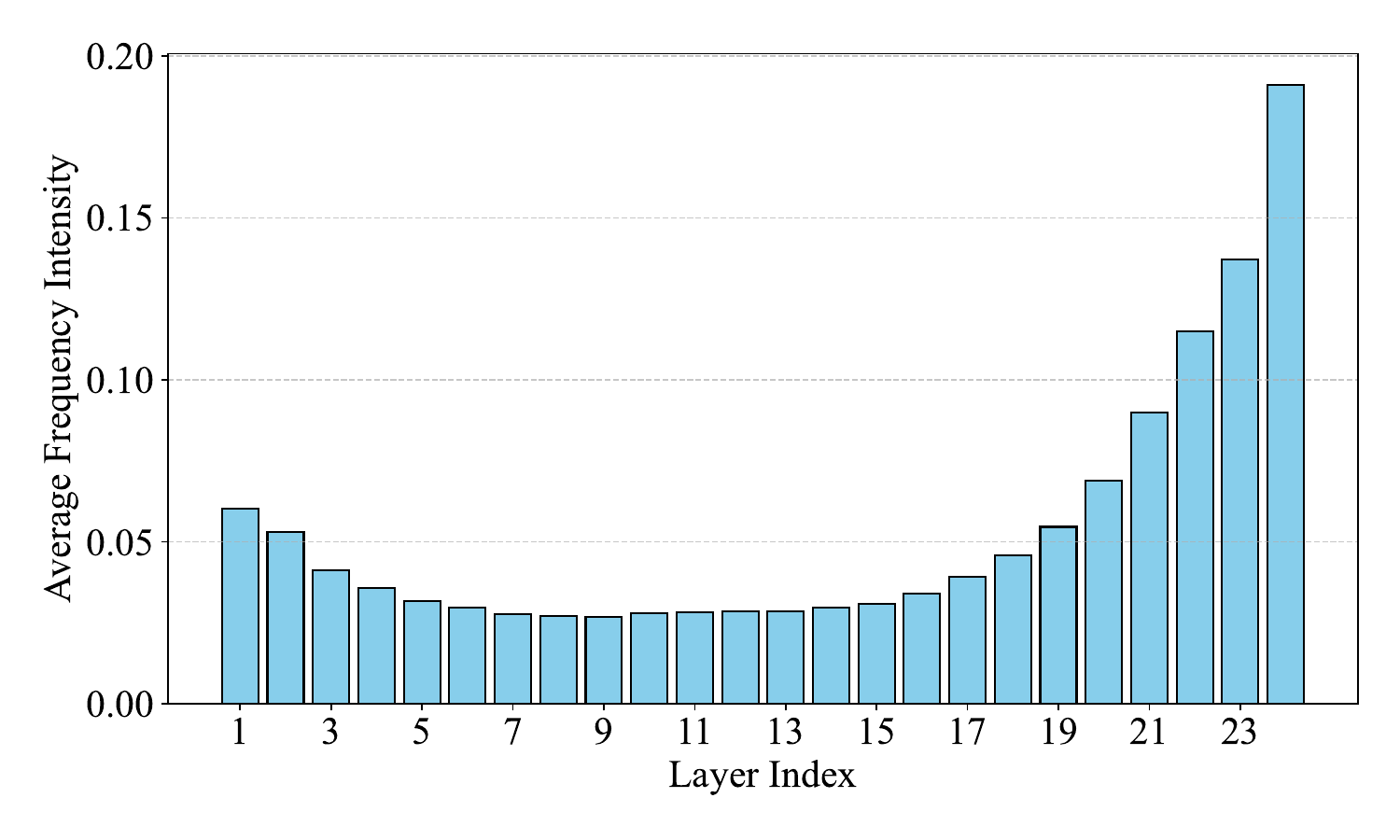}
			\label{CaiT_S24}
		}
	\end{minipage}
	\begin{minipage}[b]{\linewidth}
		\subfigure[DeiT-T without distillation.]{
			\includegraphics[width=0.46\linewidth]{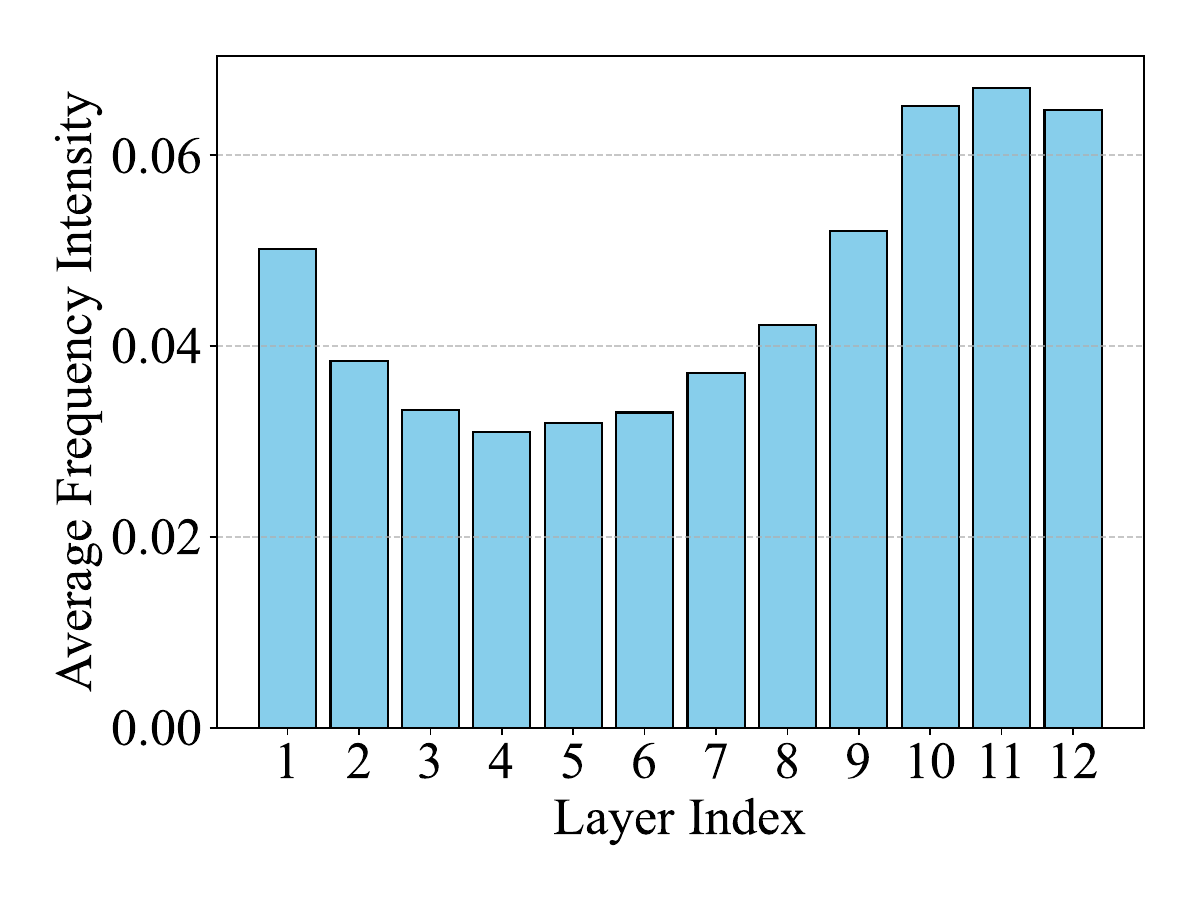}
			\label{cait_oo}
		}
		\subfigure[DeiT-T by SpectralKD.]{
			\includegraphics[width=0.46\linewidth]{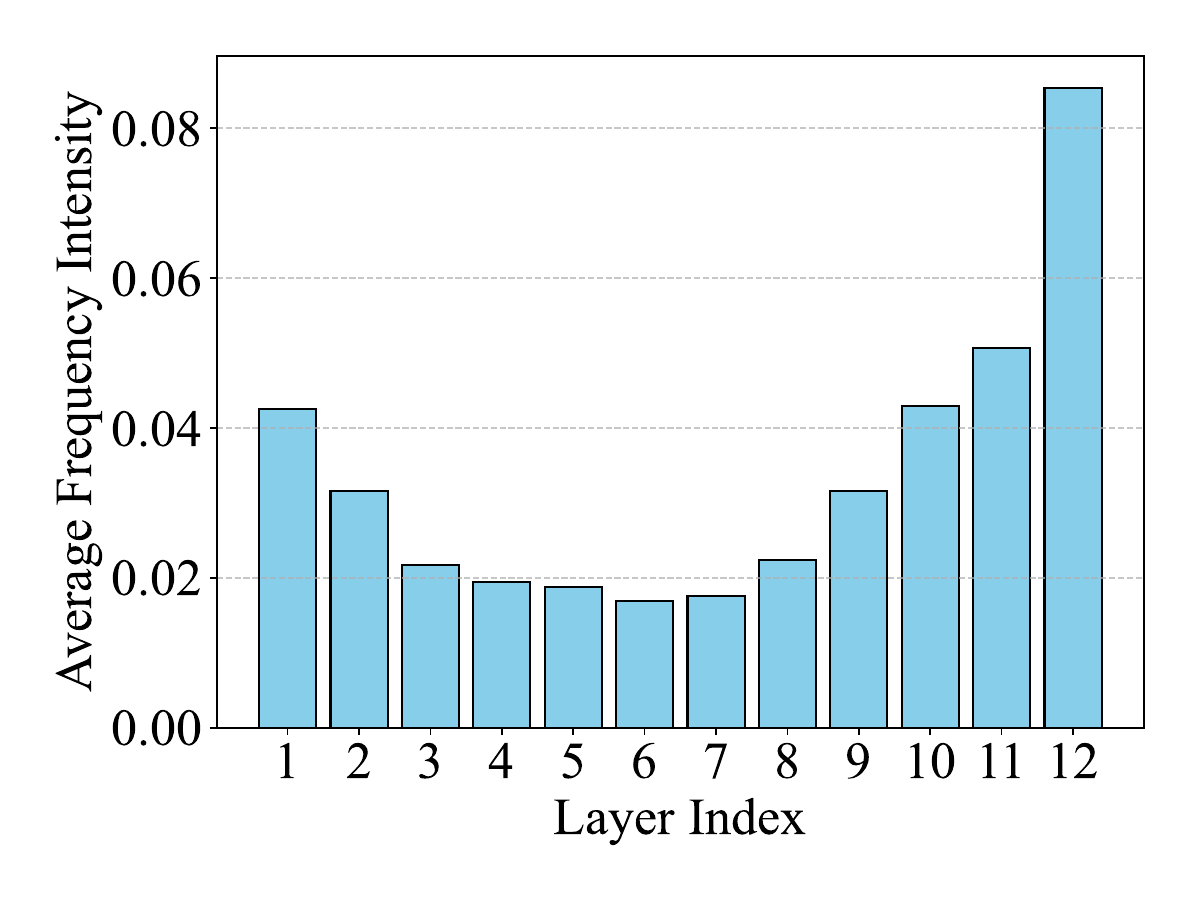}
			\label{cait_fas}
		}
	\end{minipage}
	\caption{Model-wise frequency intensity analysis $L(\mathbf{X})$ (Equation~\ref{model_intensity}), plotted across the Transformer depth for three models: (a) CaiT-S24 (teacher), (b) DeiT-Tiny without distillation, and (c) DeiT-Tiny distilled by our SpectralKD. SpectralKD clearly shifts the student's intensities in different layers closer to those of the teacher, especially in the most information-rich layers in the early and final few layers. Both CaiT-S24 and baseline DeiT-Tiny checkpoints are taken from the \emph{timm} library\cite{rw2019timm}.}
	\label{fig:Layer_wise_intensities}
\end{figure}

\section{Introduction}
\label{Intro}

Benefiting from attention mechanisms, Transformer models \cite{vaswani2017attention} have achieved remarkable success across diverse applications \cite{devlin2018bert, brown2020language, dosovitskiy2020image, han2021transformer, liu2021swin, han2022survey, xu2024self, li2024transformer, bar-shalom2024subgraphormer}. However, their substantial computational cost remains a significant challenge. KD \cite{hinton2015distilling}, an effective model compression technique, can accelerate inference and reduce resource requirements \cite{choudhary2020comprehensive, pham2024frequency, rao2024dual}. To develop trustworthy and efficient models, it is crucial to gain deeper insights into both the internal encoding patterns of Transformer models and the mechanisms underlying KD.

Recently, researchers have made significant efforts \cite{abnar2020quantifying, raghu2021vision, chefer2021transformer, yeh2023attentionviz, zimmermann2024scale, zeng2024peeling} to gain deeper insights into the attention mechanisms. Better interpretability \cite{pan2021ia, caron2021emerging, Yu_2023_CVPR} can be leveraged to improve the efficiency of ViTs. Meanwhile, numerous theoretical analyses \cite{phuong2019towards, menon2021statistical, chandrasegaran2022revisiting, allenzhu2023understandingensemble} have also been conducted to reveal the internal mechanisms of KD. However, a unified theoretical framework that encompasses both ViTs and KD remains an open research challenge.

In this paper, we propose SpectralKD, a novel unified analytical framework that enhances our understanding of ViTs and optimizes the KD process via spectral analysis. Our key contributions are summarized as follows:
\begin{itemize}
	
	\item \textbf{Model-wise analysis.} By analyzing intermediate feature maps in the frequency domain, our spectral analysis unveils a characteristic U-shaped frequency pattern in CaiT layers: early and late few layers capture richer spectral information, whereas middle layers encode lower-intensity frequencies. This model-wise observation provides guidelines for layer selection in KD when dealing with uniform Transformers.
	
	\item \textbf{Layer-wise insight.} Surprisingly, both hierarchical (Swin) and uniform (CaiT) transformer architectures exhibit similar encoding patterns, offering a potential explanation for strong generalization capabilities of ViTs. This layer-wise finding also suggests practical strategies for feature alignment in KD.
	
	\item \textbf{A simple, parameter-free KD strategy.} Building on the above insights, we propose a straightforward, parameter-free KD approach. Despite its simplicity, our method benefits from the interpretability provided by spectral analysis and achieves state-of-the-art (SOTA) performance.
	
	\item \textbf{Distillation dynamics.} Post-training analysis demonstrates that SpectralKD induces global changes in the student's spectral patterns, even in layers not explicitly aligned. This result opens up a new area we term ``distillation dynamics", thus paving the way for more transparent and interpretable KD.
	
\end{itemize}

\begin{figure}[ht]
	\vskip 0.2in
	\begin{center}
		\centerline{\includegraphics[width=\columnwidth]{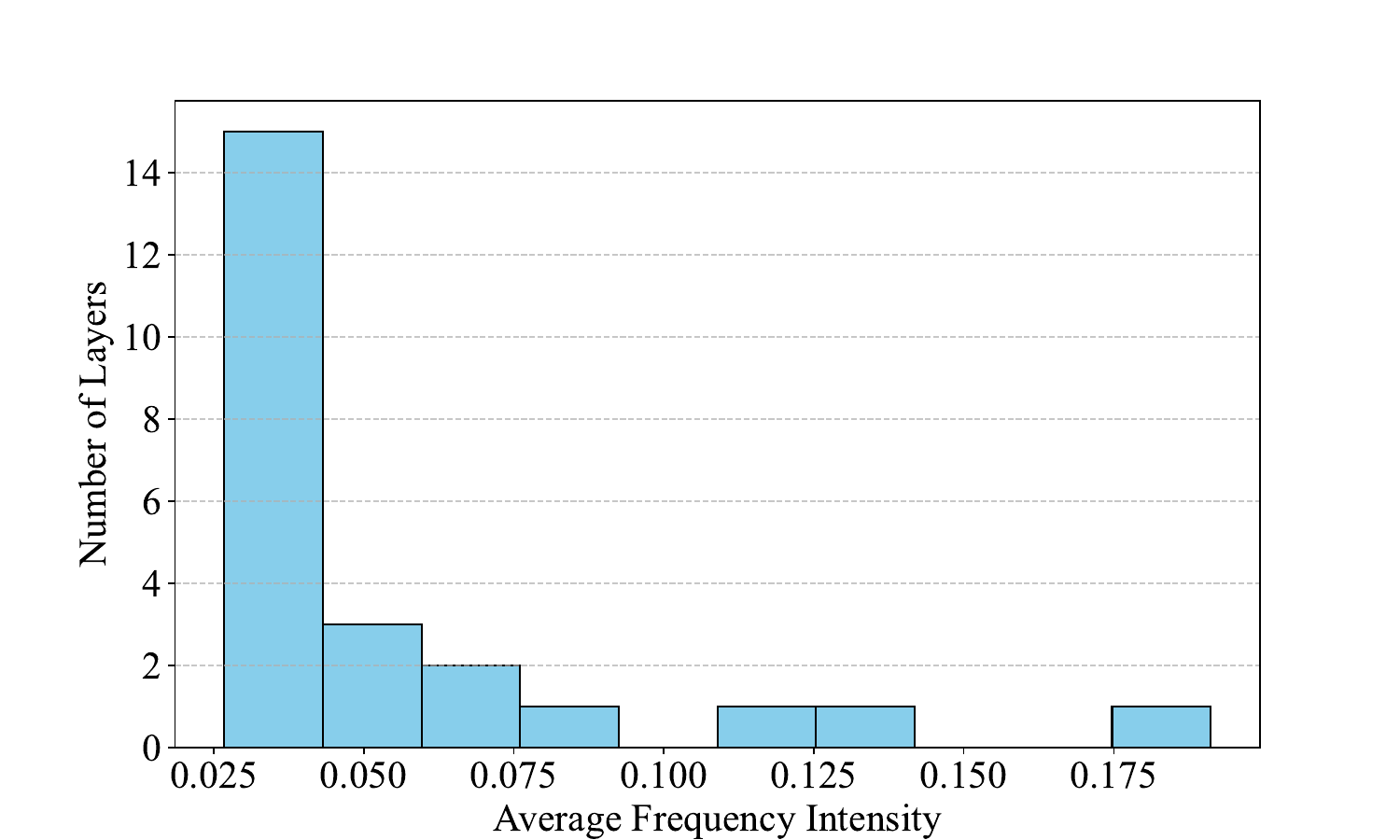}}
		\caption{Histogram of the model-wise frequency intensities $L(\mathbf{X})$ (Equation~\ref{model_intensity}) for all $24$ layers of CaiT-S24. The distribution is heavily skewed: most layers have relatively low spectral intensity, while only a small number exhibit substantially higher values. These \emph{peaks} point to potentially critical layers for knowledge distillation.}
		\label{fig:Histogram}
	\end{center}
	\vskip -0.2in
\end{figure}

\section{Spectral Analysis} 
\label{Spectral-Analysis}

In this section, we introduce SpectralKD, a unified analytical framework for ViTs and KD. Our framework provides quantitative insights into the information flow across network layers and unveils the underlying encoding patterns of ViTs. Unlike prior approaches relying primarily on empirical observations, SpectralKD offers a more theoretically grounded method to analyze and distill ViTs.

\begin{figure}[t]
	\centering
	\begin{minipage}[b]{\linewidth}
		\subfigure[Layer $1$.]{
			\includegraphics[width=0.295\linewidth]{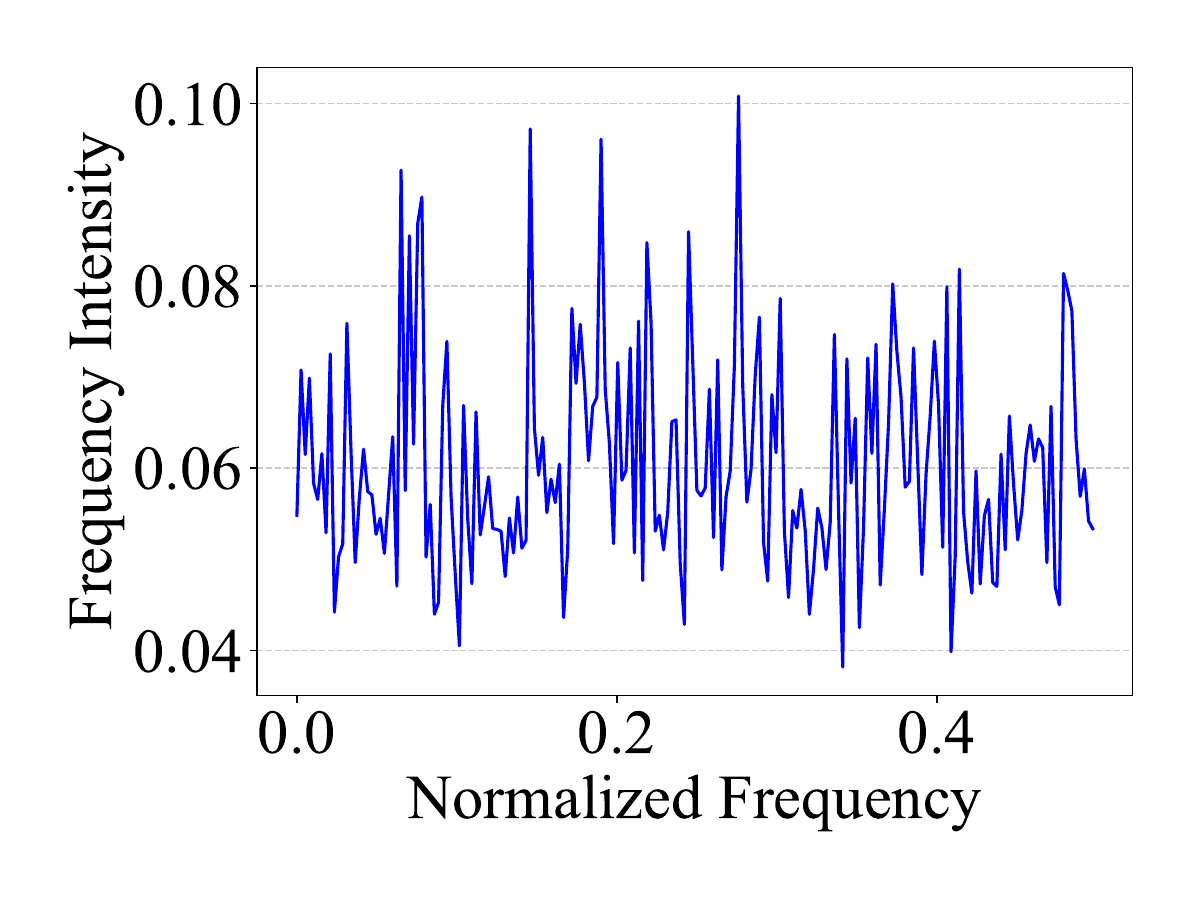}
			\label{Layer_1}
		}
		\subfigure[Layer $2$.]{
			\includegraphics[width=0.295\linewidth]{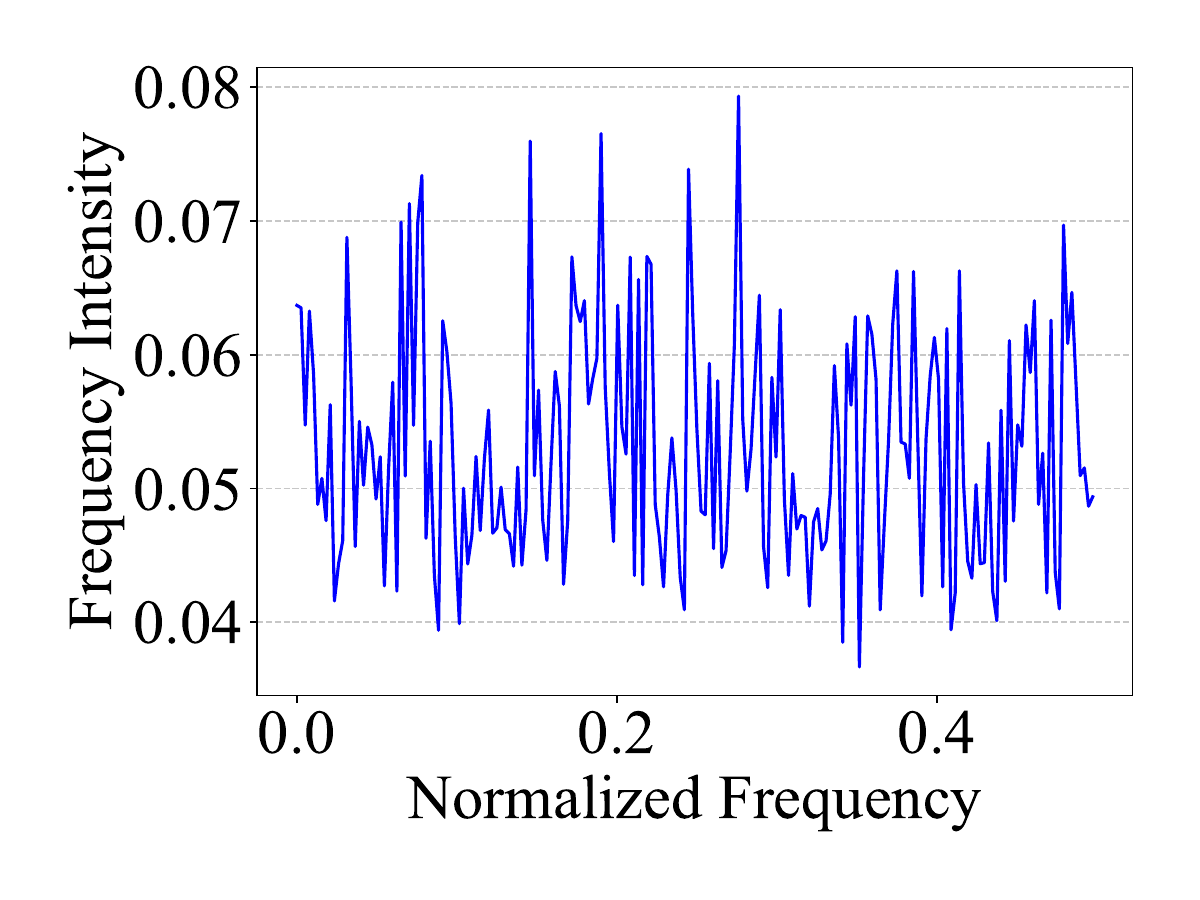}
			\label{Layer_2}
		}
		\subfigure[Layer $12$.]{
			\includegraphics[width=0.295\linewidth]{Frequency_Intensity_11_th_layer}
			\label{Layer_12}
		}
	\end{minipage}
	\begin{minipage}[b]{\linewidth}
		\subfigure[Layer $13$.]{
			\includegraphics[width=0.295\linewidth]{Frequency_Intensity_12_th_layer}
			\label{Layer_13}
		}
		\subfigure[Layer $23$.]{
			\includegraphics[width=0.295\linewidth]{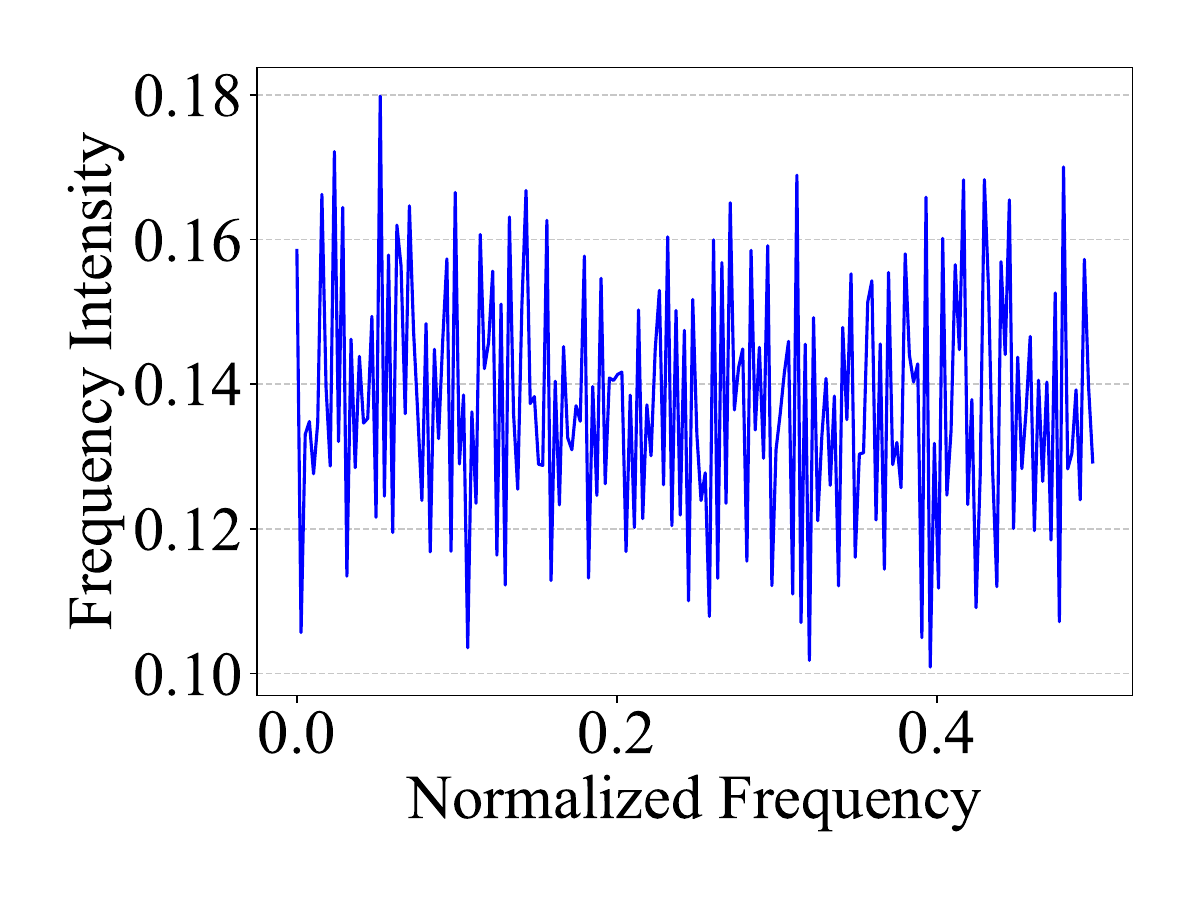}
			\label{Layer_23}
		}
		\subfigure[Layer $24$.]{
			\includegraphics[width=0.295\linewidth]{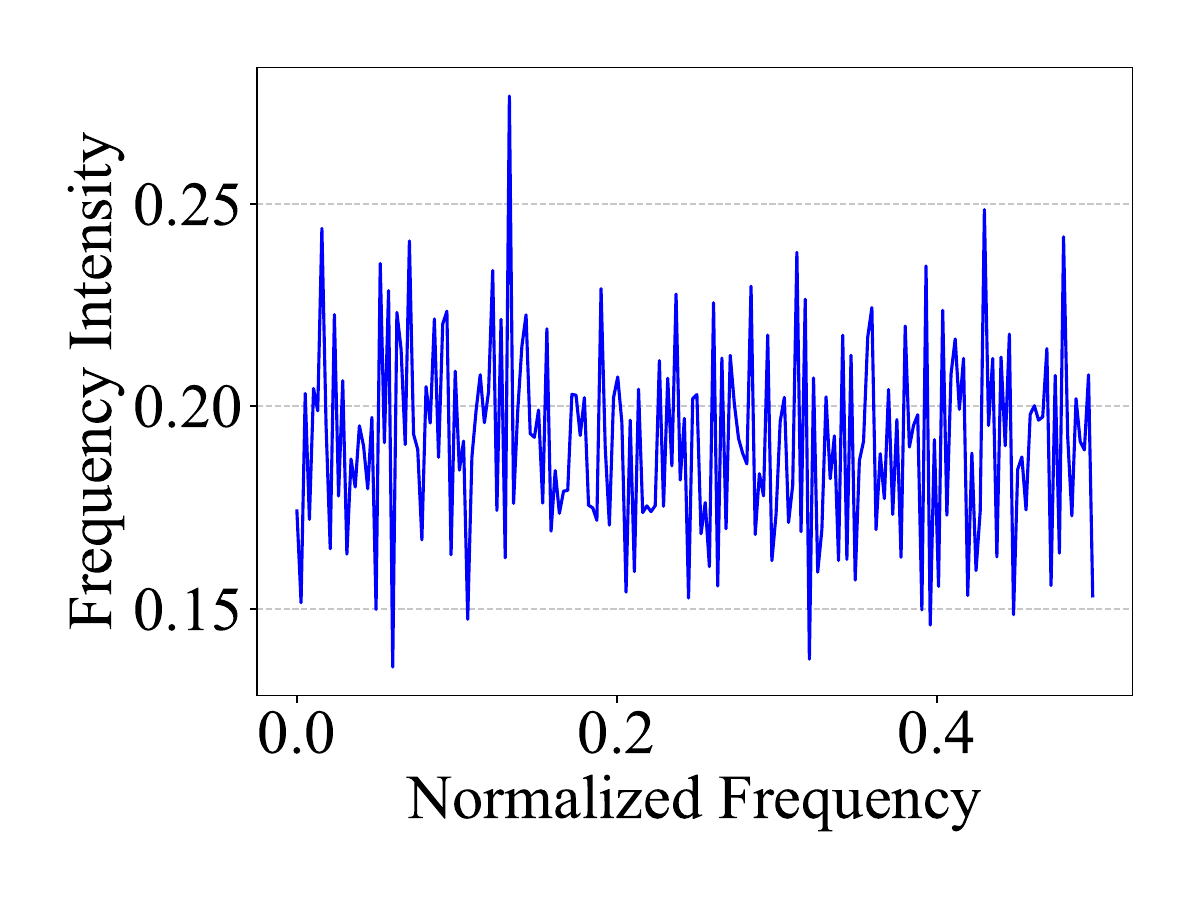}
			\label{Layer_24}
		}
	\end{minipage}
	\caption{Layer-wise spectral intensity distributions $\mathbf{S}(\mathbf{X})$ (Equation~\ref{channel_intensity}) for representative layers of CaiT-S24. The visualization reveals distinct encoding patterns across network depths.  \textbf{Early layers (a-b)} exhibit approximately uniform intensities across frequencies. \textbf{Middle layers (c-d)} show a marked decay from low to high frequency. \textbf{Final layers (e-f)} once again become relatively uniform but at distinctly higher overall intensities.}
	\label{fig:spectral_distribution}
\end{figure}

\begin{figure}[t]
	\centering
	\begin{minipage}[b]{\linewidth}
		\subfigure[Stage $1$.]{
			\includegraphics[width=0.44\linewidth]{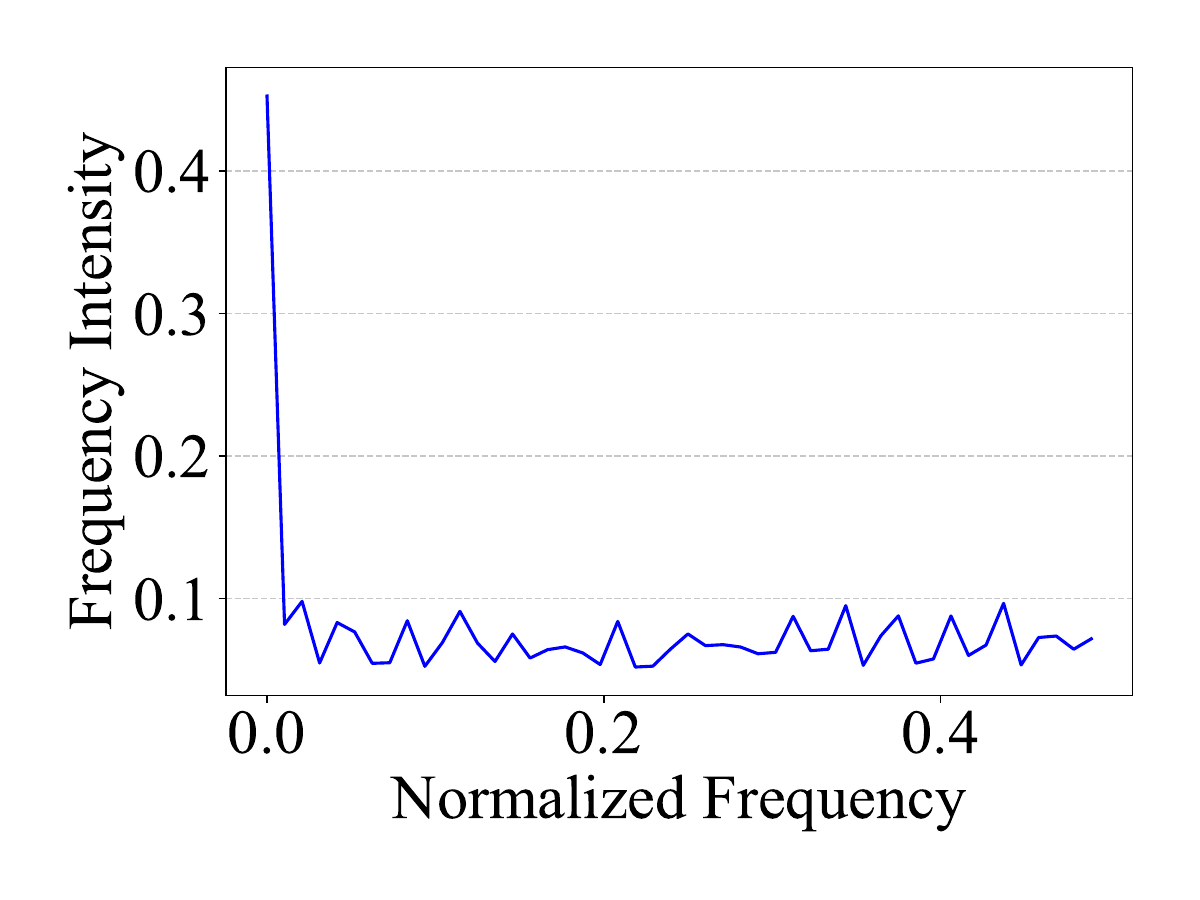}
			\label{Stage_1}
		}
		\subfigure[Stage $2$.]{
			\includegraphics[width=0.44\linewidth]{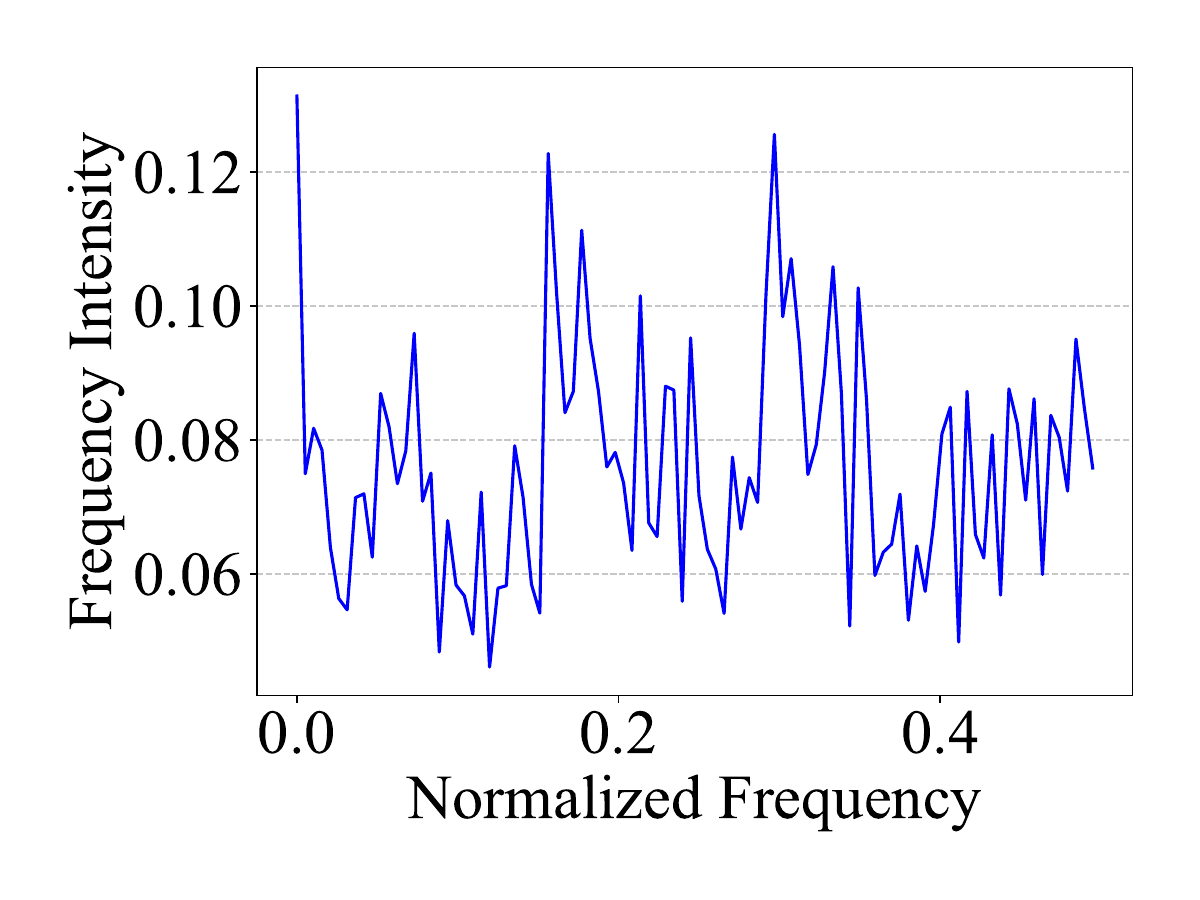}
			\label{Stage_2}
		}
	\end{minipage}
	\begin{minipage}[b]{\linewidth}
		\subfigure[Stage $3$.]{
			\includegraphics[width=0.44\linewidth]{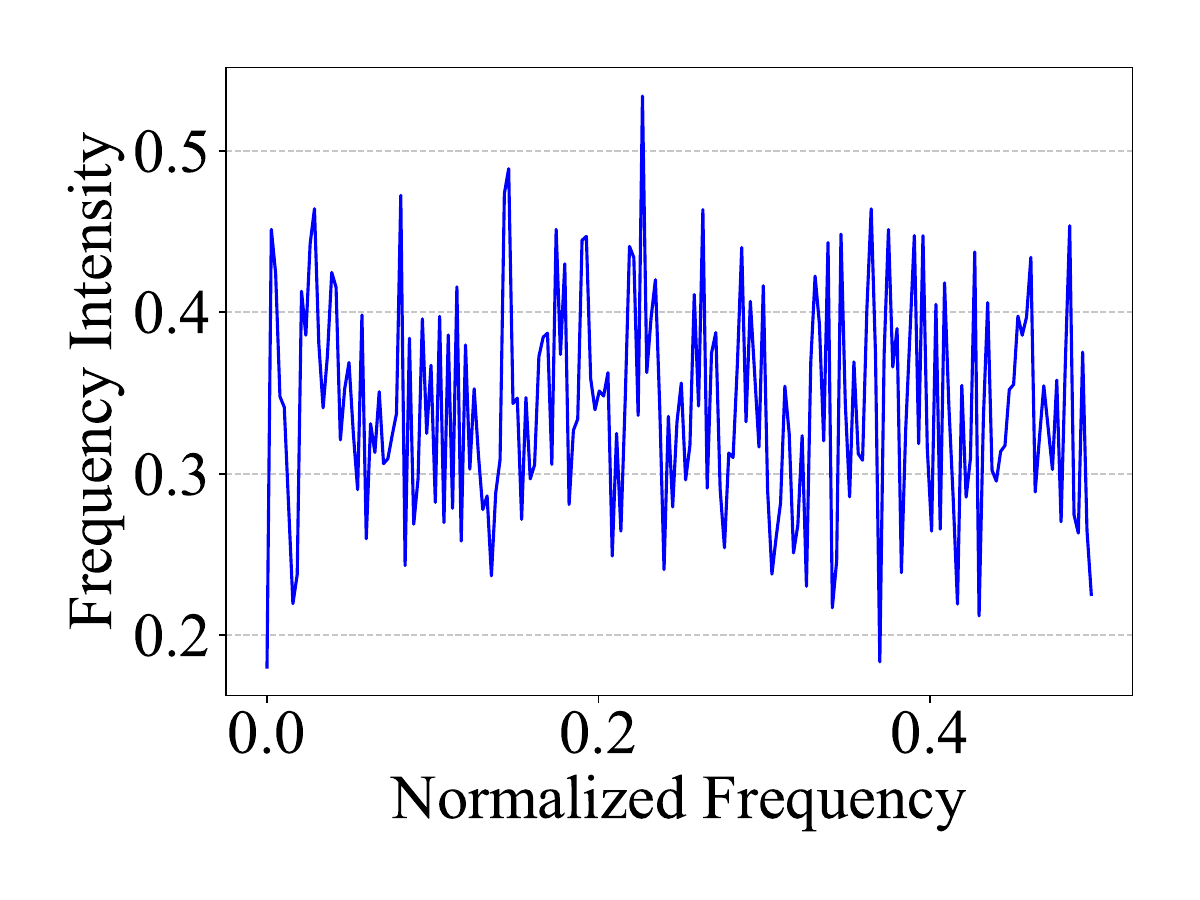}
			\label{Stage_3}
		}
		\subfigure[Stage $4$.]{
			\includegraphics[width=0.44\linewidth]{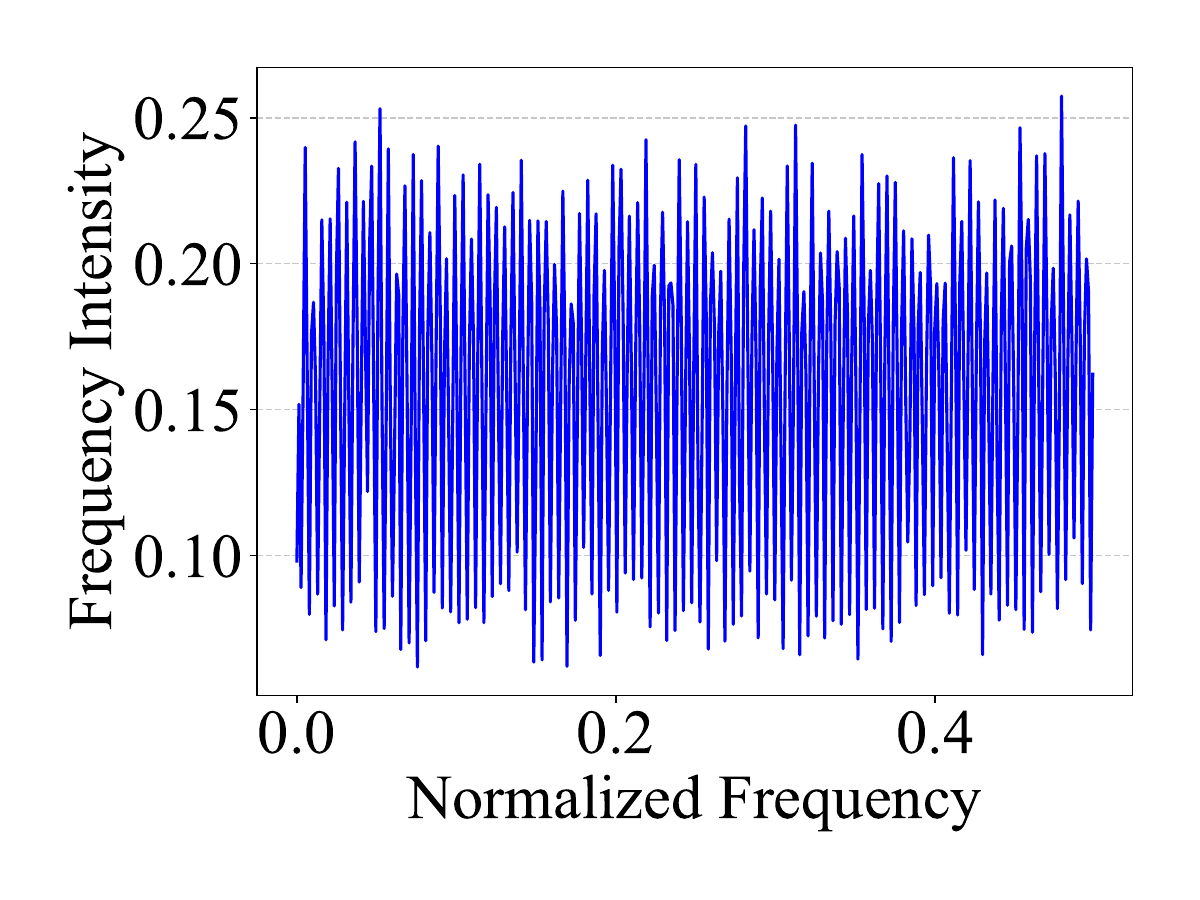}
			\label{Stage_4}
		}
	\end{minipage}
	\caption{Stage-wise spectral intensity distributions $\mathbf{S}(\mathbf{X})$ (Equation~\ref{channel_intensity}) for Swin-Small. Despite the architectural contrasts with CaiT, the same trend emerges across its four stages. \textbf{Early stages (a-b)} show a decay from low to high frequency. \textbf{Later stages (c-d)}  exhibit increasingly uniform and higher-intensity representations. This similarity suggests a shared spectral encoding strategy in deep Transformer models. \textbf{Note}: \emph{Each stage of Swin Transformer comprises multiple attention blocks, mapping to several layers in a uniform ViT model like CaiT.}}
	\label{fig:swin}
\end{figure}

\subsection{Analysis Method}

Consider a batch of intermediate feature maps $\mathbf{X} \in \mathbb{R}^{B \times C \times H \times W}$ from a particular layer, where $B$ is the batch size, $C$ is the number of channels, and $H$ and $W$ denote the spatial height and width, respectively. We apply a one-dimensional Fast Fourier Transform (FFT) along the channel dimension to map these real-valued feature maps into the complex domain $\mathbb{C}$. Formally,  
\begin{equation}  
	\mathcal{F}(\mathbf{X}) = \mathrm{FFT}(\mathbf{X}),  
\end{equation}  
where $\mathrm{FFT}$ indicates the $1$D FFT applied independently for each spatial location $(b, h,w)$. This operation involves performing $B \times H \times W$ separate $1$D FFTs over vectors of length $C$.

After obtaining the frequency domain representation $\mathcal{F}(\mathbf{X})\in \mathbb{C}^{B \times C \times H \times W}$ of the feature maps, we compute the magnitude of each complex entry as follows:  
\begin{equation}  
	\mathbf{A}(\mathbf{X}) = \sqrt{\mathrm{Re}^{2}(\mathcal{F}(\mathbf{X})) + \mathrm{Im}^{2}(\mathcal{F}(\mathbf{X}))},  
\end{equation}  
where $\mathrm{Re}(\mathcal{F}(\mathbf{X}))$ and $\mathrm{Im}(\mathcal{F}(\mathbf{X}))$ represent the real and imaginary components of $\mathcal{F}(\mathbf{X})$, respectively. This yields a real-valued tensor $\mathbf{A}(\mathbf{X})\in\mathbb{R}^{B \times C \times H \times W}$.

Next, we average $\mathbf{A}(\mathbf{X})$ over the batch dimension $B$ and spatial dimensions $H$ and $W$:

\begin{equation} \label{channel_intensity}  
	\mathbf{S}(\mathbf{X})= \frac{1}{B \times H \times W}\sum_{b=1}^{B} \sum_{h=1}^{H} \sum_{w=1}^{W} \mathbf{A}(\mathbf{X}).  
\end{equation}  
obtaining $\mathbf{S}(\mathbf{X}) \in \mathbb{R}^{C}$. Each element of $\mathbf{S}(\mathbf{X})$ corresponds to the average spectral intensity of a particular frequency component. Hence, $\mathbf{S}(\mathbf{X})$ offers a \emph{layer-wise} view into the encoding patterns of neural networks in the spectral domain.

To quantify the overall frequency intensity of a single layer, we further average $\mathbf{S}(\mathbf{X})$ over the channel dimension:  
\begin{equation}  
	\ell(\mathbf{X}) = \frac{1}{C} \sum_{c=1}^{C} \mathbf{S}(\mathbf{X}).
\end{equation}  
The scalar $\ell(\mathbf{X})$ thus captures the aggregate spectral intensity of feature map of a single layer or stage in the model. Higher $\ell(\mathbf{X})$ values typically indicate layers encoding more complex or diverse features, making them strong candidates for transfer in KD.

Finally, by computing $\ell(\mathbf{X})$ across multiple layers or stages, we obtain
\begin{equation}  
	L(\mathbf{X}) = \{\ell^{(1)}(\mathbf{X}), \ell^{(2)}(\mathbf{X}), \dots, \ell^{(n)}(\mathbf{X})\}, \label{model_intensity}  
\end{equation}  
where $n$ denotes the total number of layers or stages under consideration. The collection $L(\mathbf{X})$ provides a \emph{model-wise} perspective on how information complexity evolves with network depth.

In summary, our spectral analysis provides both \emph{model-wise} and \emph{layer-wise} insights into the flow of information in models.  This dual perspective provides new analytical tools for machine learning models.

\subsection{Analysis of ViTs} \label{analysis-vit}  

In this section, we show how ViTs can be analyzed through SpectralKD, and then optimize KD.

\subsubsection{Model-Wise Analysis} \label{model-wise} 

To investigate how information flows through ViTs, SpectralKD framework is applied to CaiT-S24 \cite{touvron2021cait}, a model containing 24 Transformer layers. Figures~\ref{CaiT_S24} and~\ref{fig:Histogram} present two complementary visualizations of how frequency intensity evolves across layers, as computed by Equation~\ref{model_intensity}.

The results in these figures highlight two main findings:
\begin{enumerate}
	\item \textbf{U-Shaped Intensity Curve.} As seen in Figure \ref{CaiT_S24}, the average frequency intensities exhibit a  U-shaped trend. The first and  final few layers have substantially higher intensities, suggesting that they encode more information-rich representations. Conversely, the middle layers demonstrate much lower intensity values, implying their role may focus more on feature extraction and transformation than on encoding new information.
	\item \textbf{Skewed Distribution of Spectral Magnitude.}  The histogram in Figure~\ref{fig:Histogram} confirms that only a few layers possess notably higher intensities, while the majority cluster around lower magnitude values. This skewed distribution suggests that certain peak layers could be especially beneficial in KD, as they appear to concentrate more complex or salient features.
\end{enumerate}

\subsubsection{Layer-Wise Analysis} \label{layer-wise}

We next investigate the \emph{layer-wise} spectral distribution (computed according to Equation~\ref{channel_intensity}) within individual feature maps at various network depths. Figure~\ref{fig:spectral_distribution} presents the frequency distributions of six representative layers in CaiT-S24, revealing three main patterns:

\begin{enumerate}  
	\item \textbf{Early layers (e.g., Figure~\ref{Layer_1}, \ref{Layer_2}).} These layers show relatively uniform intensity distributions across the frequency bands but maintain moderate overall intensities. This suggest that fine-grained (high-frequency) information is preserved early on, likely to capture detailed features from the input.  
	\item \textbf{Middle layers (e.g., Figure~\ref{Layer_12}, \ref{Layer_13}, layers~12-13).} In contrast to the early layers, the middle layers show a pronounced decay from low to high frequencies. This pattern implies that the network at this depth transforms the features into more abstract and smoother representations. Such an observation may yield new insights into how ViTs achieve strong generalization.  
	\item \textbf{Final layers (e.g., Figure~\ref{Layer_23}, \ref{Layer_24}).} In the last few layers, the spectral distributions reverts to a more uniform distribution but with significantly higher intensities. This pattern indicates a strong utilization of the channel dimension to encode both high- and low-frequency information, which is beneficial for fine-grained classification.  
\end{enumerate}

A complete visualization of spectral distributions across all layers of CaiT is available in Appendix~\ref{Appendix-A}.

\subsubsection{Cross-Architecture Analysis}  \label{cross-architecture}  

To broaden our analysis, we further study the Swin-Small \cite{liu2021swin} model, which adopts a hierarchical design quite distinct from CaiT. Each stage of Swin Transformer comprises multiple attention blocks, mapping to several layers in a uniform model like CaiT. Thus, the first stage of Swin Transformer may contain enough attention blocks parallel to middle layers of CaiT.

As shown in Figure~\ref{fig:swin}, the frequency distributions across its four stages display a surprisingly similar pattern to CaiT, highlighting a convergent encoding strategy of both architectures. Similar to the middle layers of CaiT, the early stages of Swin Transformer show a decay from low to high frequency. The later stage exhibits an almost uniform distribution in spectral domain at relatively high magnitudes, which is also similar to the final layers of CaiT.

Interestingly, patterns in both later stages and final layers differ from typical natural signals (where intensity tends to decay at higher frequencies). This suggests that deep networks develop their own artificial signal characteristics, actively encoding final representations based on abstract features in the middle layers or early stages rather than merely preserving natural image statistics.

\subsubsection{Implications for KD} \label{guide}

A deeper understanding of both the \emph{model-wise} flow of information and the \emph{layer-wise} encoding patterns in ViTs is essential for designing an effective knowledge transfer strategy. In feature-based KD, two primary challenges arise:

\begin{enumerate}
	\item Identifying the best layers for distillation from an architectural perspective, i.e., which teacher layers should guide the student.
	\item Aligning features across the channel dimension within intermediate representations, so that knowledge from the teacher truly informs the student’s internal encoding.
\end{enumerate}

While hierarchical architectures such as Swin Transformer naturally suggest distillation points at stage boundaries, uniform Transformers (such as ViT \cite{dosovitskiy2020image}, CaiT \cite{touvron2021cait}, and DeiT \cite{touvron2021deit}) consist of repeated blocks that appear structurally identical, making it less obvious which layers are most valuable for KD.

In conventional signal processing \cite{gonzales1987digital}, having strong spectral intensity does not necessarily imply information-rich representations. However, our \emph{model-wise} and \emph{layer-wise} analyses reveal a noteworthy ViT-specific phenomenon: layers with higher aggregate spectral intensity $\ell(\mathbf{X})$ tend to exhibit more uniform frequency energy distributions. Empirically, these layers encode a broader range of frequencies, from low-frequency global patterns to high-frequency local details. We hypothesize that regularization mechanisms (e.g., Layer Normalization) in ViTs may help flatten frequency energy curves, thus unlocking more multi-band capacity in certain layers.

Accordingly, we adopt spectral intensity as a practical proxy for information richness in ViTs. Our results consistently show that layers with high  $\ell(\mathbf{X})$ are well-suited for distillation. Based on these observations, we arrive at two concrete guidelines:
\begin{enumerate}
	\item \textbf{For uniform Transformers.} Early and final few layers in uniform Transformers like CaiT have notably higher intensities, indicating they carry richer multi-frequency content. Distilling from these layers is often more beneficial than from the middle ones.
	\item \textbf{For alignment of feature maps.} Since the most informative layers also show strong utilization of all frequency bands, feature alignment in distillation should preserve this multi-frequency encoding. Aligning only a subset of channel may miss this crucial encoding pattern.
\end{enumerate}

In the next section, we propose a simple distillation strategy built on the above practical guidelines, and then in the Section \ref{Experiments}, we present empirical evaluations that confirm these conclusions. 

\section{Frequency Alignment for KD} \label{frequency-alignment}

Motivated by the above insights, we introduce a frequency alignment distillation method that explicitly aligns the spectral characteristics of student and teacher networks. 

\subsection{Method Overview}

Let $\mathbf{F}_s \in \mathbb{R}^{B \times C_s \times H \times W}$ and $\mathbf{F}_t \in \mathbb{R}^{B \times C_t \times H \times W}$ denote the intermediate feature maps of student and teacher models from particular layer or stage, respectively, where $B$ is the batch size, $C_s$ and $C_t$ denote the number of channels, and $H, W$ represent the spatial dimensions. 

Our approach first aligns the channel dimensions of these feature maps and then performs a $2$D FFT over their spatial dimensions \footnote{In earlier spectral analysis (Section~\ref{Spectral-Analysis}), we use $1$D FFT across channels to characterize frequency intensities. Here, we use $2$D FFT across spatial dimensions to align global and local features, which is more natural for data like images.} before computing the distillation loss. By applying a $2$D Fourier transform to their feature maps, we capture and align both low-frequency (global) and high-frequency (fine-grained) details, thus enabling richer knowledge transfer.

\subsection{Alignment Strategy}
\label{alignment-strategy}

\textbf{Channel Dimension Alignment.} To adapt different channel numbers between student and teacher, we apply $3$D adaptive average pooling:
\begin{align}
	\mathbf{F}_s  & = \text{AdaptiveAvgPool}(\mathbf{F}_s) & \text{if } C_s > C_t, \\
	\mathbf{F}_t  & = \text{AdaptiveAvgPool}(\mathbf{F}_t) & \text{if } C_s < C_t.
\end{align}
After this operation, both feature maps are in the shape of $\mathbb{R}^{B \times C \times H \times W}$, where $C = \min(C_s, C_t)$. We choose adaptive average pooling over linear projections or additional attention layers for two reasons: (1) it preserves the spatial structure of features while adjusting channel dimension, and (2) it remains simple enough to allow clear interpretability of the distillation process. This design follows our earlier analysis (Section~\ref{analysis-vit}), which finds that Transformer networks tend to leverage all channels for information encoding.

\textbf{Fast Fourier Transform.} We then apply the $2$D real-valued Fast Fourier Transform (RFFT2) along the spatial dimensions of each channel in the aligned feature maps:
\begin{align}
	\mathcal{F}(\mathbf{F}_s) = \text{RFFT2}(\mathbf{F}_s),\\
	\mathcal{F}(\mathbf{F}_t) = \text{RFFT2}(\mathbf{F}_t).
\end{align}
To simplify computation and loss definition, we separate the real and imaginary parts and then stack them along a new dimension:
\begin{align}
	\mathcal{F}_{\text{real}}(\mathbf{F}_s) &  = \text{Re}(\mathcal{F}(\mathbf{F}_s)), \\
	\mathcal{F}_{\text{imag}}(\mathbf{F}_s) & = \text{Im}(\mathcal{F}(\mathbf{F}_s)), \\
	\mathcal{F}_{\text{stack}}(\mathbf{F}_s) &  = \text{Stack}(\mathcal{F}_{\text{real}}(\mathbf{F}_s), \mathcal{F}_{\text{imag}}(\mathbf{F}_s)).
\end{align}
The same operations are applied to the feature maps of teacher.

\subsection{Loss Function}

\textbf{Frequency Alignment Loss.} We use the Mean Squared Error (MSE) loss to measure the difference between the student and teacher representations in the frequency domain:
\begin{equation}
	\mathcal{L}_{\text{FFT}} = \text{MSE}(\mathcal{F}_{\text{stack}}(\mathbf{F}_s) - \mathcal{F}_{\text{stack}}(\mathbf{F}_t)).
\end{equation}
This penalizes discrepancies across both low- and high-frequency components, ensuring that the student learns to capture global structure (low frequency) as well as fine details (high frequency).

\textbf{Training Objective.} To harness both conventional KD loss and our proposed frequency alignment, we combine $\mathcal{L}_{\text{FFT}}$ with the standard KD loss \cite{hinton2015distilling}:
\begin{equation}
	\begin{aligned}
		\mathcal{L}_{\text{KD}} &= (1-\alpha) \mathcal{L}_{\text{CE}}(f_{s}(\mathbf{x}), y) \\
		&+ \alpha T^{2} \mathcal{L}_{\text{KL}}\left(\frac{f_{s}(\mathbf{x})}{T}, \frac{f_{t}(\mathbf{x})}{T}\right),
	\end{aligned}
\end{equation}
where $\mathcal{L}_{\text{CE}}$ is the cross-entropy loss between the student predictions $f_{s}(\mathbf{x})$ and ground-truth labels $y$, and $\mathcal{L}_{\text{KL}}$ is the Kullback-Leibler divergence. The temperature $T$ smooths the logits of teacher, and $\alpha$ balances the two terms.

We then form the total loss:
\begin{equation}
	\mathcal{L}_{\text{Total}} = \mathcal{L}_{\text{KD}} + \beta \mathcal{L}_{\text{FFT}},
\end{equation}
where $\beta$ controls how heavily the frequency alignment term influences training. This objective encourages the student to learn both decision boundaries (through standard KD) and internal frequency representations (through $\mathcal{L}_{\text{FFT}}$), ultimately yielding a more robust and comprehensive knowledge transfer.

\begin{table*}[t]
	\caption{Classification accuracies on ImageNet-1K for DeiT-Tiny and DeiT-Small.}
	\label{DeiT-cls}
	\vskip 0.15in
	\centering
	\begin{center}
		\begin{small}
			\begin{sc}
				\begin{tabular}{lccccr}
					\toprule
					\textbf{Distillation method} & \textbf{Teacher} & \textbf{Params} & \textbf{Top-1 (\%)} & \textbf{Student} & \textbf{Top-1 (\%)} \\ \midrule
					-                            & -                & -               & -                   & DeiT-Tiny (5M)   & 72.2                \\ \hline
					Hard  \cite{touvron2021deit}                       & RegNetY-16GF     & 84M             & 82.9                & DeiT-Tiny (5M)   & 74.5                \\
					DearKD \cite{chen2022dearkd}             & RegNetY-16GF     & 84M             & 82.9                & DeiT-Tiny (5M)   & 74.8                \\
					USKD \cite{yang2023knowledge}               & RegNetY-16GF     & 84M             & 82.9                & DeiT-Tiny (5M)   & 75.0                \\
					SRD \cite{miles2024srd}                & RegNetY-16GF     & 84M             & 82.9                & DeiT-Tiny (5M)   & 77.2                \\ \hline
					Hard \cite{touvron2021deit}                        & CaiT-S24         & 47M             & 83.4                & DeiT-Tiny (5M)   & 74.5                \\
					Manifold \cite{hao2022manifold}        & CaiT-S24         & 47M             & 83.4                & DeiT-Tiny (5M)   & 76.5                \\
					MaskedKD \cite{son2025maskedkd}           & CaiT-S24         & 47M             & 83.4                & DeiT-Tiny (5M)   & 75.9                \\
					\textbf{SpectralKD (Ours)}             & CaiT-S24         & 47M             & 83.4                & DeiT-Tiny (5M)   & \textbf{77.4}                \\ \hline \hline
					-                            & -                & -               & -                   & DeiT-Small (22M) & 79.9                \\
					Hard \cite{touvron2021deit}                        & RegNetY-16GF     & 84M             & 82.9                & DeiT-Small (22M) & 81.2                \\
					DearKD \cite{chen2022dearkd}             & RegNetY-16GF     & 84M             & 82.9                & DeiT-Small (22M) & 81.5                \\
					USKD \cite{yang2023knowledge}               & RegNetY-16GF     & 84M             & 82.9                & DeiT-Small (22M) & 80.8                \\
					SRD \cite{miles2024srd}               & RegNetY-16GF     & 84M             & 82.9                & DeiT-Small (22M) & 82.1                \\ \hline
					Hard \cite{touvron2021deit}                        & CaiT-S24         & 47M             & 83.4                & DeiT-Small (22M) & 81.3                \\
					Manifold \cite{hao2022manifold}        & CaiT-S24         & 47M             & 83.4                & DeiT-Small (22M) & \textbf{82.2}                \\
					\textbf{SpectralKD (Ours)}             & CaiT-S24         & 47M             & 83.4                & DeiT-Small (22M) & \textbf{82.2}   \\            
					
					\bottomrule
				\end{tabular}
			\end{sc}
		\end{small}
	\end{center}
	\vskip -0.1in
\end{table*}

\begin{table*}[t]
	\caption{Classification accuracies on ImageNet-1K for Swin-Tiny. $\ddagger$: Pretrained on  ImageNet-22K.}
	\label{Swin-cls}
	\vskip 0.15in
	\centering
	\begin{center}
		\begin{small}
			\begin{sc}
				\begin{tabular}{lccccr}
					\toprule
					\textbf{Distillation method} & \textbf{Teacher} & \textbf{Params} & \textbf{Top-1 (\%)} & \textbf{Student} & \textbf{Top-1 (\%)} \\\midrule
					-                            & -                & -                   & -                   & Swin-Tiny (29M)  & 81.3                \\ \hline
					KD  \cite{hinton2015distilling}                         & Swin-Large $\ddagger$       & 197M                & 86.3                & Swin-Tiny (29M)  & 81.5                \\
					RKD \cite{park2019rkd}                & Swin-Large $\ddagger$       & 197M                & 86.3                & Swin-Tiny (29M)  & 81.2                \\
					SRRL \cite{yang2021srrl}               & Swin-Large $\ddagger$        & 197M                & 86.3                & Swin-Tiny (29M)  & 81.5                \\
					DIST \cite{huang2022dist}            & Swin-Large  $\ddagger$      & 197M                & 86.3                & Swin-Tiny (29M)  & 82.3                \\
					ScaleKD \cite{fan2024scalekd}         & Swin-Large $\ddagger$       & 197M                & 86.3                & Swin-Tiny (29M)  & \textbf{83.8}                \\ \hline
					Manifold \cite{hao2022manifold}        & Swin-Small       & 50M                 & 83.2                & Swin-Tiny (29M)  & 82.2                \\
					\textbf{SpectralKD (Ours)}             & Swin-Small       & 50M                 & 83.2                & Swin-Tiny (29M)  & \textbf{82.7}       \\        
					\bottomrule
				\end{tabular}
			\end{sc}
		\end{small}
	\end{center}
	\vskip -0.1in
\end{table*}

\section{Experiments} \label{Experiments}

We demonstrate the effectiveness of SpectralKD through extensive experiments on image classification task. Our results show that our SpectralKD leads to SOTA performance. This section details our experimental setup, presents comprehensive results, and validates our approach through ablation studies.

\subsection{Experimental Setup}

\textbf{Dataset and Models.} We conduct experiments on the ImageNet-1k dataset \cite{deng2009imagenet}, which comprises 1.28M training images and 50K validation images across 1,000 classes. We use multiple vision transformer architectures for our evaluations.

\textbf{Implementation Details.} All experiments are conducted on $2$ NVIDIA RTX 4090D GPUs with a batch size of $256$. Training DeiT-Tiny requires approximately $184$ GPU hours. Our implementation is in PyTorch, and we build upon the \emph{timm} library \cite{rw2019timm} for model architectures and pretrained weights.

\textbf{Student Networks.} We evaluate several vision transformer variants as student models, all trained exclusively on ImageNet-1k: (1) DeiT-Tiny and DeiT-Small, both trained from scratch following the DeiT settings \cite{touvron2021deit}, and (2) Swin-Tiny, also trained from scratch based on the original Swin Transformer settings \cite{liu2021swin}.

\textbf{Teacher Networks.} We employ CaiT and Swin-Small architectures as teacher models, both pretrained solely on ImageNet-1k (i.e., without ImageNet-22k pretraining). We obtain the teacher checkpoints from the timm library.

\textbf{Hyperparameters.} Unless otherwise stated, we use a distillation temperature of $1$,  $\alpha  = 0.9$, and $\beta = 0.2$ for DeiT experiments, while $\beta = 0.05$ is used for Swin Transformer experiments.

\subsection{Results and Analysis}

Table~\ref{DeiT-cls} compares our SpectralKD derived method with SOTA KD approaches on ImageNet-1k, using CaiT-S24 ($47$M parameters) as the teacher. Guided by our \emph{model-wise} and \emph{layer-wise} spectral analysis insights of ViTs (Section~\ref{analysis-vit}), we apply distillation to the first two and final six layers of both teacher and student. We train DeiT-Tiny for $400$ epochs and DeiT-Small for $500$ epochs under the DeiT training protocol \cite{touvron2021deit}, setting stochastic depth rate to $0$.

\textbf{DeiT-Tiny Results.} With DeiT-Tiny ($5$M parameters) as the student model, SpectralKD achieves SOTA performance with a top-1 accuracy of $77.4\%$, an absolute improvement of $5.2\%$ over the $72.2\%$ baseline. It also outperforms the standard hard distillation baseline of $74.5\%$, demonstrating more efficient knowledge transfer.

\textbf{DeiT-Small Results.} For the larger DeiT-Small student ($22$M parameters), SpectralKD delivers similarly strong gains. It improves the baseline accuracy from $79.9\%$ to $82.2\%$, outperforming the baseline by $2.3\%$ and the hard distillation ($81.3\%$) by $0.9\%$ in absolute terms. The consistent performance across model scales underscores the effectiveness of our frequency-based distillation strategy for knowledge transfer.

\textbf{Swin Transformer Results.} We further evaluate SpectralKD on hierarchical transformer architectures. Using Swin-Small ($50$M parameters) as the teacher model and Swin-Tiny ($29$M parameters, $81.3\%$ baseline) as the student model, SpectralKD achieves $82.7\%$ top-1 accuracy (Table \ref{Swin-cls}), improving on the baseline by $1.4\%$. This result indicates that SpectralKD effectively transfers knowledge across hierarchical transformer architectures with different attention mechanisms and feature hierarchies compared to DeiT models.

\subsection{Ablation Study}

To further validate the effectiveness of SpectralKD, we conduct ablation studies on ImageNet-1k with DeiT-Tiny as the student model. 

Table~\ref{Ablation-study} presents the results of different KD configurations. When combining SpectralKD with soft KD, the student model achieves the best performance with $77.4\%$ top-1 accuracy, surpassing the baseline by $5.2\%$. The superior performance of the full model demonstrates that SpectralKD complements traditional knowledge distillation by capturing valuable spectral information not explicitly present in either hard or soft predictions.

\begin{table}[h]
	\caption{Ablation study results on ImageNet-1K. DeiT-Tiny serves as the student model.}
	\label{Ablation-study}
	\vskip 0.15in
	\begin{center}
		\begin{small}
			\begin{sc}
				\begin{tabular}{lccr}
					\toprule
					\textbf{Method}                 & \textbf{Top-1 (\%)} & \textbf{$\Delta$ (\%)} \\ \midrule
					w/o KD                & 72.2                & -                      \\
					Hard KD             & 74.5                & +2.3                   \\
					Soft KD             & 76.2                & +4.0                   \\
					Soft KD + SpectralKD & \textbf{77.4}                & \textbf{+5.2}                  \\
					\bottomrule
				\end{tabular}
			\end{sc}
		\end{small}
	\end{center}
	\vskip -0.1in
\end{table}

\begin{table*}[t]
	\caption{Layer matching strategies and corresponding performance on ImageNet-1K between CaiT-S24 (teacher) and DeiT-Tiny (student). $\dagger$: Spectral analysis-based layer selection as described in Section~\ref{Spectral-Analysis}.}
	\label{Layer-ablation}
	\vskip 0.15in
	\begin{center}
		\begin{small}
			\begin{sc}
				\begin{tabular}{lccr}
					\toprule
					\textbf{Matching Strategy} & \textbf{Teacher Layers $\mathcal{T}$} & \textbf{Student Layers $\mathcal{S}$} & \textbf{Top-1 (\%)} \\ \midrule
					Early-Late                 & \{1, 2, 3, 4, 21, 22, 23, 24\}        & \{1, 2, 3, 4, 9, 10, 11, 12 \}        & 77.2                \\
					Middle                     & \{4, 5, 6, 7, 8, 9, 10, 11\}          & \{10, 11, 12, 13, 14, 15, 16, 17\}    & 77.0                \\
					Spectral $\dagger$                  & \{1, 2, 19, 20, 21, 22, 23, 24\}      & \{1, 2, 7, 8, 9, 10, 11, 12 \}        & \textbf{77.4}   \\     
					\bottomrule
				\end{tabular}
			\end{sc}
		\end{small}
	\end{center}
	\vskip -0.1in
\end{table*}

Table~\ref{Layer-ablation} compares several layer matching strategies for CaiT-S24 (teacher) and DeiT-Tiny (student): uniform early-late layer matching, middle layer matching, and our proposed spectral-based layer matching. Aligning teacher layers $\mathcal{T} = \{1, 2, 19, 20, 21, 22, 23, 24\}$ with student layers 
$\mathcal{S} = \{1, 2, 7, 8, 9, 10, 11, 12\}$ via spectral analysis achieves the highest top-1 accuracy of $77.4\%$. This demonstrates that using non-uniform early and late few layers with high information intensity, as suggested by our SpectralKD framework, results in the most effective knowledge transfer.

\section{Distillation Dynamics} \label{distillation-dynamics}

We analyze the dynamics of knowledge transfer between teacher and student models using our SpectralKD framework, providing insights into how the teacher's encoding patterns shape the student’s internal representations. Specifically, we investigate how intermediate features evolve during distillation and how these changes correlate with performance improvements.

Recall that the spectral intensity of each Transformer layer (Equation~\eqref{model_intensity}) quantifies its information processing capacity. We compare three configurations: (1) the teacher model (CaiT-S24, $83.4\%$ top-1 accuracy), (2) the baseline student (DeiT-Tiny, $72.2\%$ top-1 accuracy) trained without distillation, and (3) the distilled student (DeiT-Tiny) trained with SpectralKD ($77.4\%$ top-1 accuracy). Figure~\ref{fig:Layer_wise_intensities} illustrates the \emph{model-wise} spectral intensity distributions for these three configurations.

\subsection{Teacher vs. Non-Distilled Baseline}

As discussed in Section~\ref{analysis-vit}, the teacher model (Figure~\ref{CaiT_S24}) exhibits a distinct U-shaped spectral pattern, characterized by high intensity in early layers, a dip in the middle, and a resurgence in the final layers. In contrast, the non-distilled baseline (Figure~\ref{cait_oo}) shows a similar but less pronounced curve. Its early and late layers exhibit weaker spectral peaks, and the transition from the middle to the final layers is less well-defined. In addition, the last few layers fluctuate in intensity, indicating inefficiencies in how the student encodes features. In other words, without distillation, the student seems to capture fewer high-level features and lacks the robust capacity to suppress irrelevant signals, a capability that the distilled student inherits, as we show next.

\subsection{Impact of SpectralKD}

When SpectralKD is applied using only the teacher’s first two and last six layers for alignment, the resulting student (Figure~\ref{cait_fas}) displays a spectral profile that more closely matches the teacher’s. The early and late few layers exhibit sharper intensity increases, mirroring the teacher's feature extraction and encoding patterns. Meanwhile, the middle layers display a streamlined intensity profile with reduced redundancy. This structural alignment likely contributes to the distilled student's improved generalization, as evidenced by its $+5.2\%$ top-1 accuracy gain over the baseline.

A noteworthy observation is that this teacher-like spectral distribution arises even though only a subset of student layers is explicitly paired with teacher layers. The remaining student layers adapt themselves to match the overall hierarchical encoding style of the teacher. In short, selectively distilling from critical layers can induce more global, teacher-like behavior throughout the entire student network.

\begin{figure}[h]
	\centering
	\begin{minipage}[b]{\linewidth}
		\subfigure[Layer $1$.]{
			\includegraphics[width=0.295\linewidth]{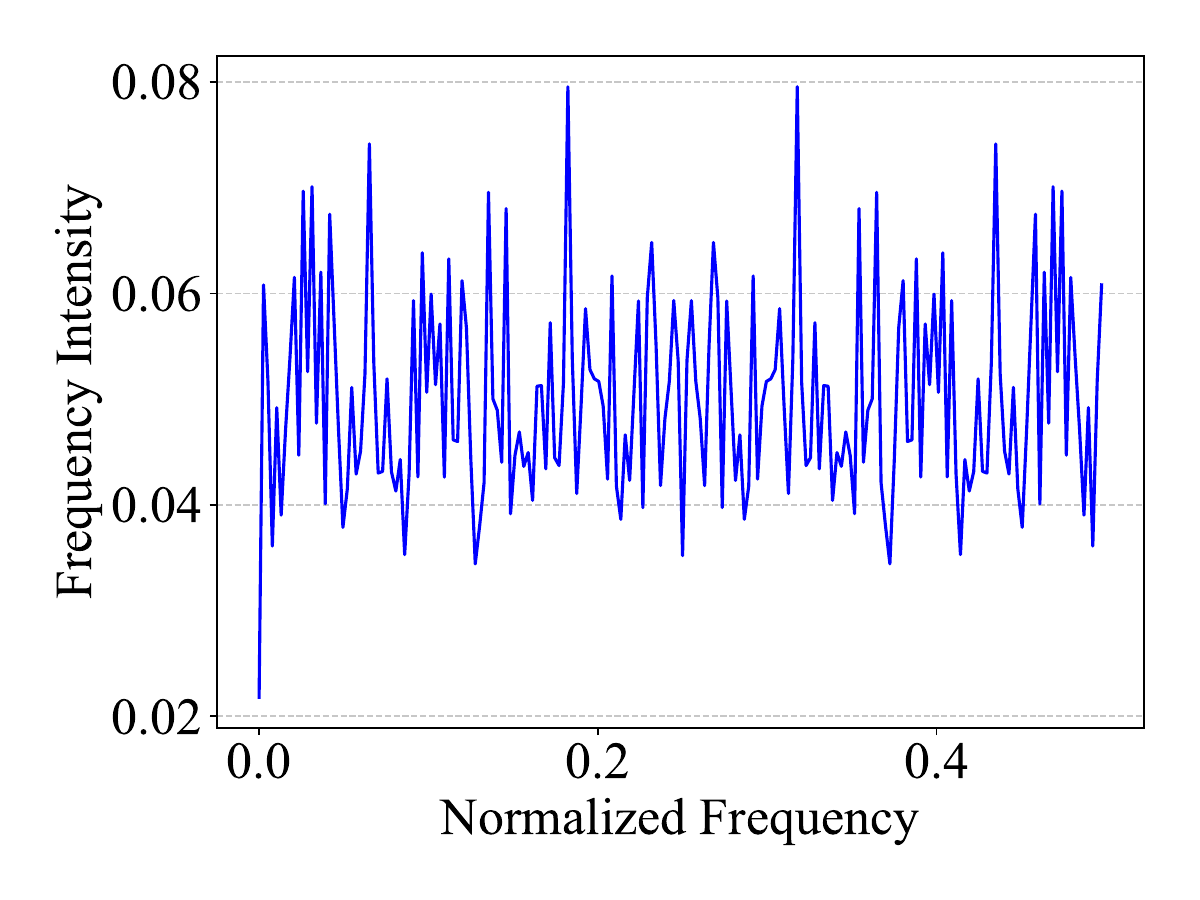}
			\label{Layer_1_oo}
		}
		\subfigure[Layer $6$.]{
			\includegraphics[width=0.295\linewidth]{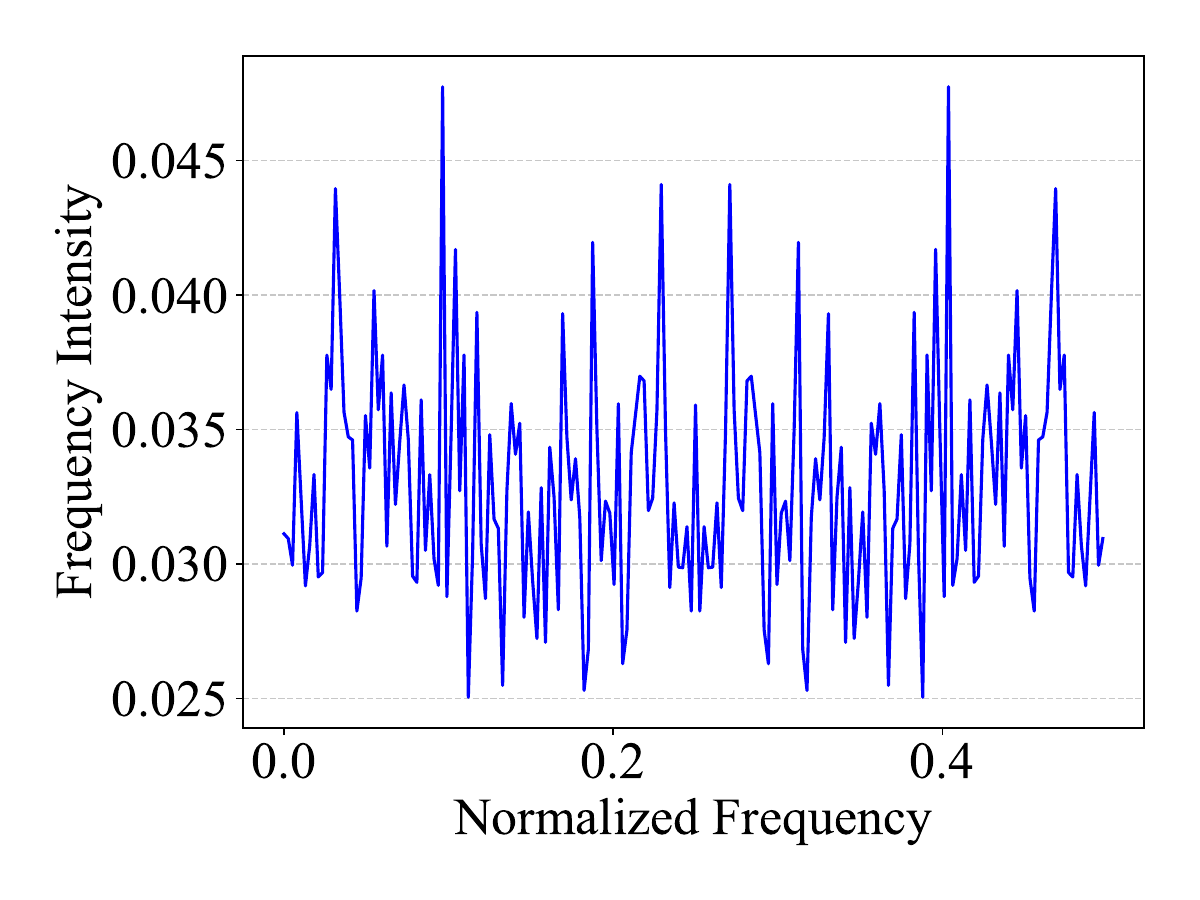}
			\label{Layer_6_oo}
		}
		\subfigure[Layer $12$.]{
			\includegraphics[width=0.295\linewidth]{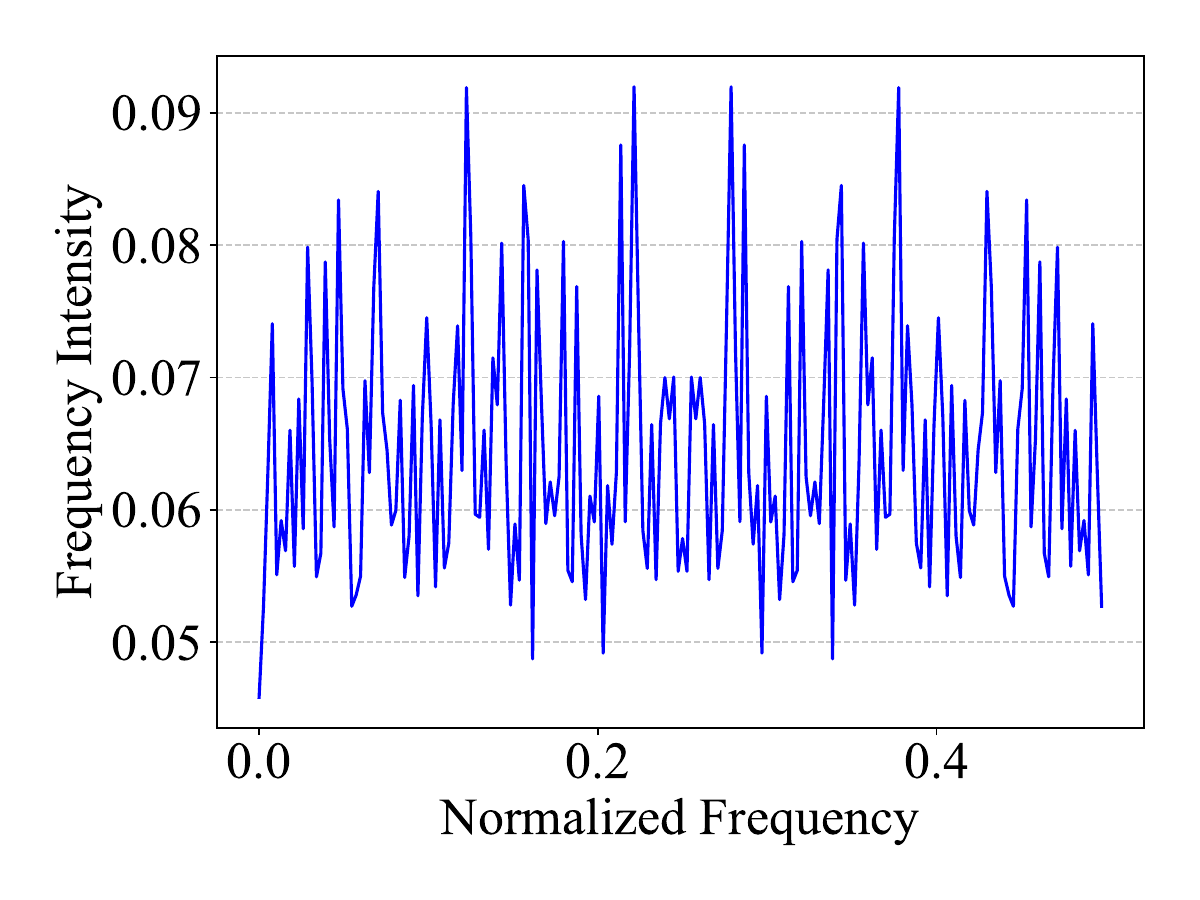}
			\label{Layer_12_oo}
		}
	\end{minipage}
	\begin{minipage}[b]{\linewidth}
		\subfigure[Layer $1$.]{
			\includegraphics[width=0.295\linewidth]{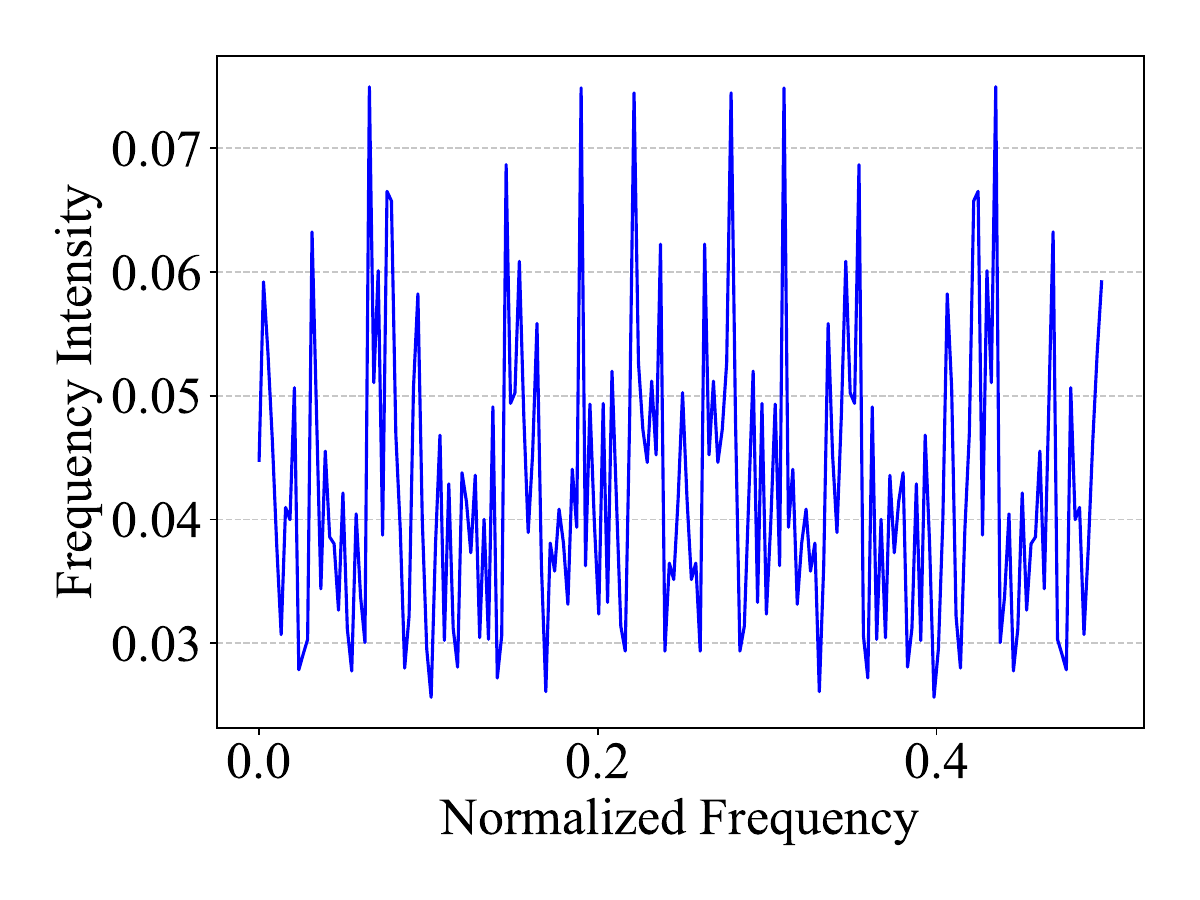}
			\label{Layer_1_dis}
		}
		\subfigure[Layer $6$.]{
			\includegraphics[width=0.295\linewidth]{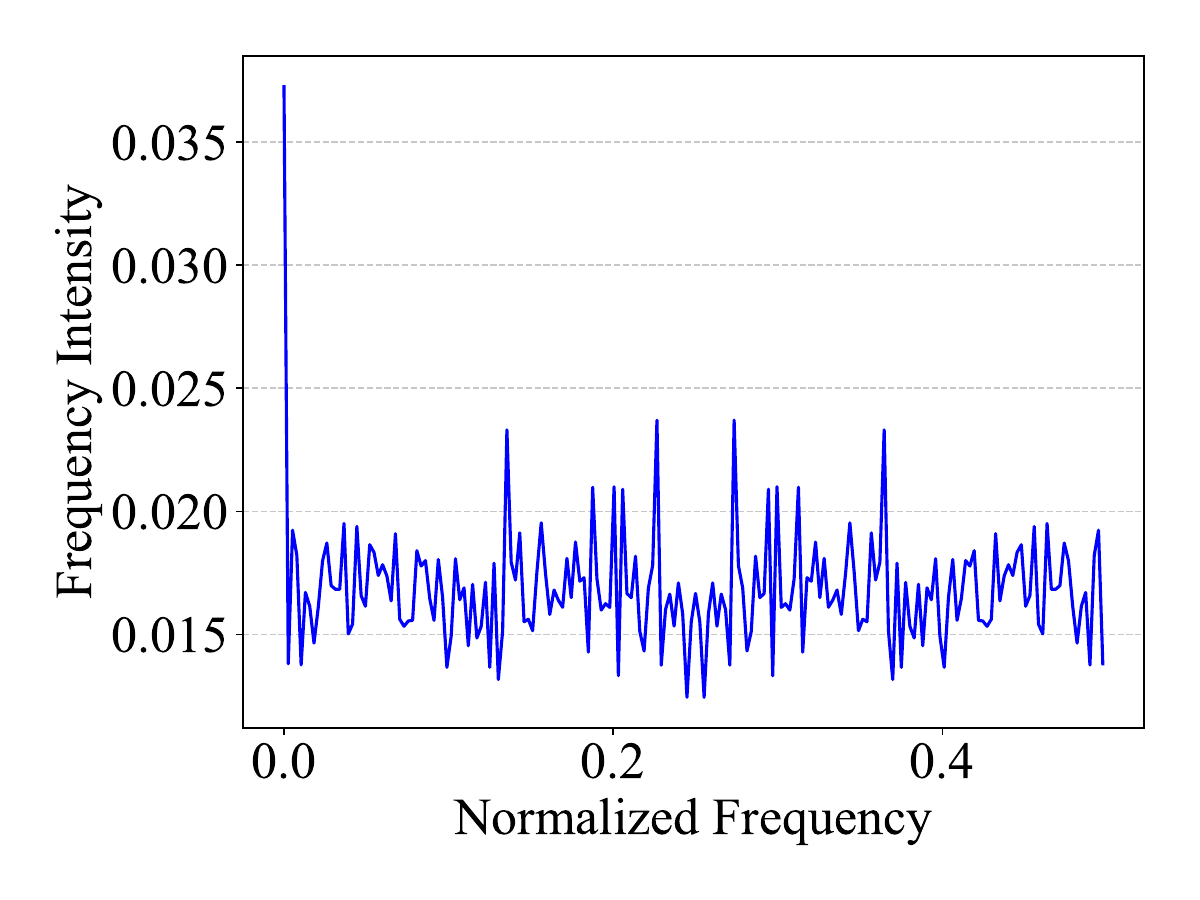}
			\label{Layer_6_dis}
		}
		\subfigure[Layer $12$.]{
			\includegraphics[width=0.295\linewidth]{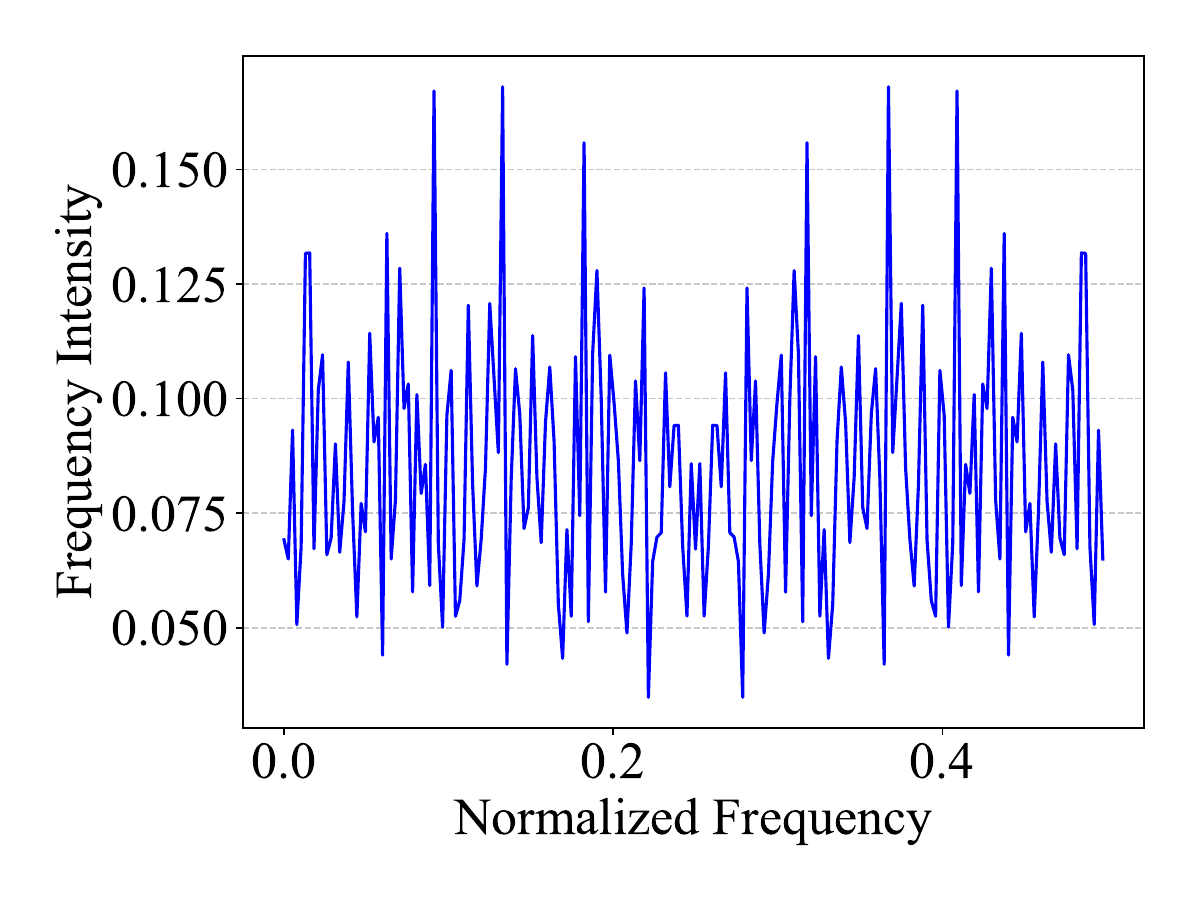}
			\label{Layer_12_dis}
		}
	\end{minipage}
	\caption{Layer-wise spectral intensity distributions $\mathbf{S}(\mathbf{X})$ (Equation~\ref{channel_intensity}) for three representative layers $(1, 6, 12)$ in the \textbf{non-distilled baseline} (top row, a–c) and the \textbf{SpectralKD-distilled student} (bottom row, d–f). Notably, the distilled student (DeiT-Tiny) exhibits teacher-like patterns (see Figure \ref{fig:spectral_distribution}) even in Layer $6$, which was not explicitly aligned during distillation. This underscores how selective alignment of early and final layers can induce broader teacher-like behavior in the student network.}
	\label{fig:dynamic_spectral_distribution}
\end{figure}

Figure~\ref{fig:dynamic_spectral_distribution} provides a more granular view by comparing spectral intensities of the non-distilled baseline (top row) and the SpectralKD-distilled student (bottom row) for Layers~$1$, $6$, and~$12$. While Layer~$1$ and Layer~$12$ are directly distilled, Layer~$6$ is not. Nevertheless, Layer~$6$ in the distilled student still exhibits partially teacher-like encoding, highlighting how early- and late-layer guidance can propagate beneficial changes across intermediate layers.

\subsection{Teacher as Implicit Feature Refiner}

Our findings suggest that, beyond compressing the model, the teacher acts as an \emph{implicit feature refiner}, passing on more noise-tolerant representations that the student uses to identify and focus on discriminative patterns. Even those middle student layers not directly distilled end up learning smoother, more abstract features than the baseline student. This observation resonates with our earlier findings (Section~\ref{analysis-vit}) that the teacher’s middle layers filter out unnecessary high-frequency details while preserving crucial semantic information.

Hence, effective KD goes beyond matching output distributions: it also involves aligning the \emph{internal} patterns that reflect how the teacher organizes information. By replicating the teacher’s hierarchical flow of features, the student gains in both robustness and accuracy. These insights not only provide more transparent KD, but also open a new research area we term ``distillation dynamics".

\section{Conclusion}

In this paper, we introduce SpectralKD, a unified framework for interpreting and distilling ViTs via spectral analysis. Our key contributions stem from both theoretical insights and practical advancements. Our analyses (Section~\ref{Spectral-Analysis}) reveal two key findings about ViTs: (1) in uniform transformer architectures, only a few early and late layers exhibit particularly high spectral intensities, guiding us toward optimal points for KD, and (2) despite their different designs, hierarchical transformers and uniform transformers share remarkably similar layer-wise encoding patterns, deriving alignment guidelines for KD.

Building on these observations, we propose a simple, parameter-free KD strategy (Section~\ref{frequency-alignment}), achieving SOTA performance on ImageNet-1K (Section~\ref{Experiments}). Post-training analysis (Section~\ref{distillation-dynamics}) demonstrates that the distilled student mirrors the teacher's spectral patterns and hierarchical information flow, underscoring the ability of KD to transfer both local and global knowledge.

These results advance our understanding of how and why KD works for ViTs, opening a new research area we term ``distillation dynamics". By bridging interpretability and practical distillation, SpectralKD provides an analytical tool for future work in model compression, transfer learning, and transparent deep learning.

\section*{Impact Statement}

SpectralKD contributes to both the interpretability and practical deployment of ViTs. From a scientific perspective, our framework offers a systematic way to analyze how ViTs encode information across layers, enabling more transparent model design and analysis. Practically, the method reduces the computational and memory footprint of large ViTs, lowering energy consumption during training and inference. This efficiency gain is critical for deploying ViTs in resource-constrained settings (e.g., mobile devices, edge computing).

The interpretability insights from SpectralKD, such as identifying critical layers and cross-architecture similarities, can guide ethical AI development by demystifying feature learning in black-box models. We encourage extending this work to diverse architectures (e.g., convolutional hybrids) and tasks (e.g., segmentation), while addressing ethical risks like bias amplification. By prioritizing efficiency, transparency, and reproducibility, SpectralKD aligns with the broader goal of democratizing robust and trustworthy AI.

\nocite{langley00}

\bibliography{example_paper}
\bibliographystyle{icml2025}

\newpage
\appendix
\onecolumn
\section{Complete Spectral Distribution Analysis Across All Layers of CaiT} \label{Appendix-A}

\begin{figure}[!h]
	\centering
	\begin{minipage}[b]{\linewidth}
		\subfigure[Layer $1$.]{
			\includegraphics[width=0.46\linewidth]{Frequency_Intensity_0_th_layer.pdf}
		}
		\subfigure[Layer $2$.]{
			\includegraphics[width=0.46\linewidth]{Frequency_Intensity_1_th_layer.pdf}
		}
	\end{minipage}
	\begin{minipage}[b]{\linewidth}
		\subfigure[Layer $3$.]{
			\includegraphics[width=0.46\linewidth]{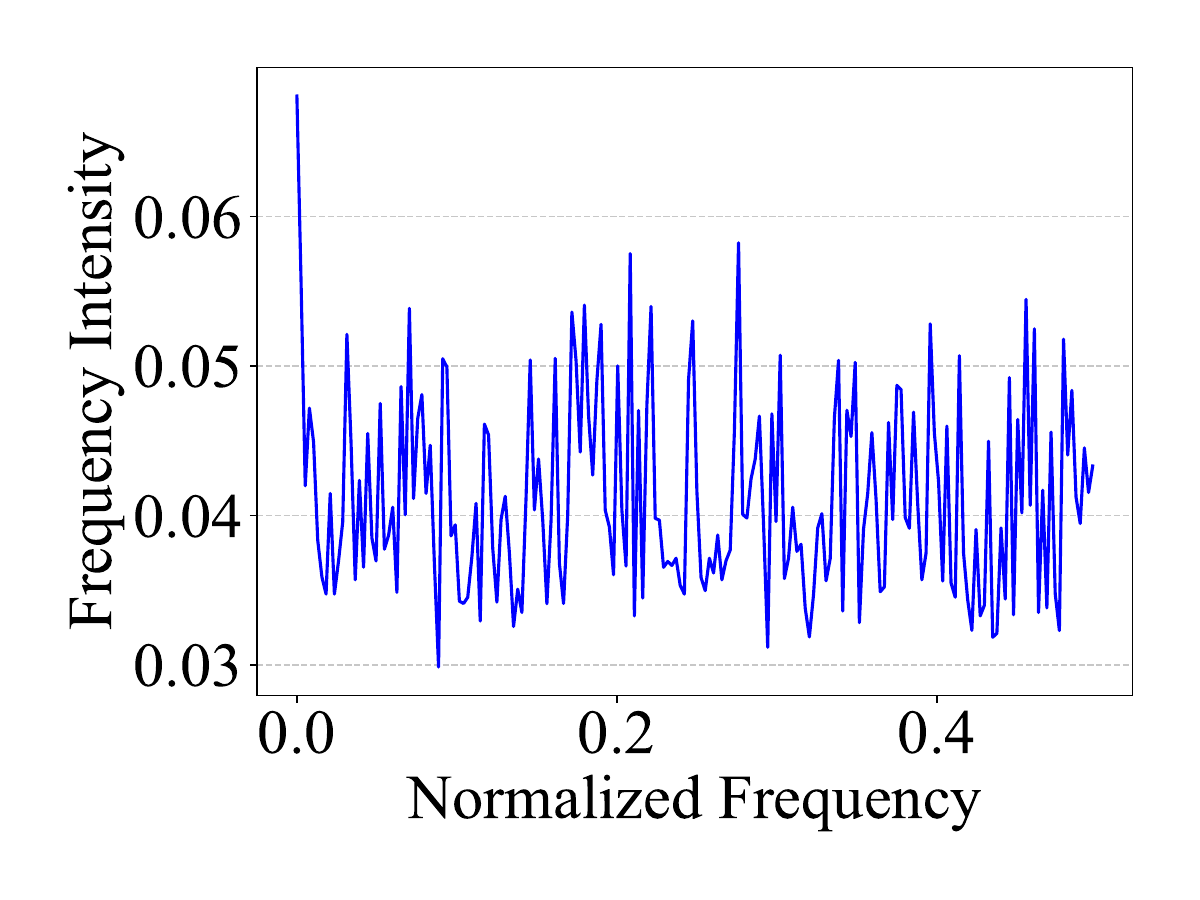}
		}
		\subfigure[Layer $4$.]{
			\includegraphics[width=0.46\linewidth]{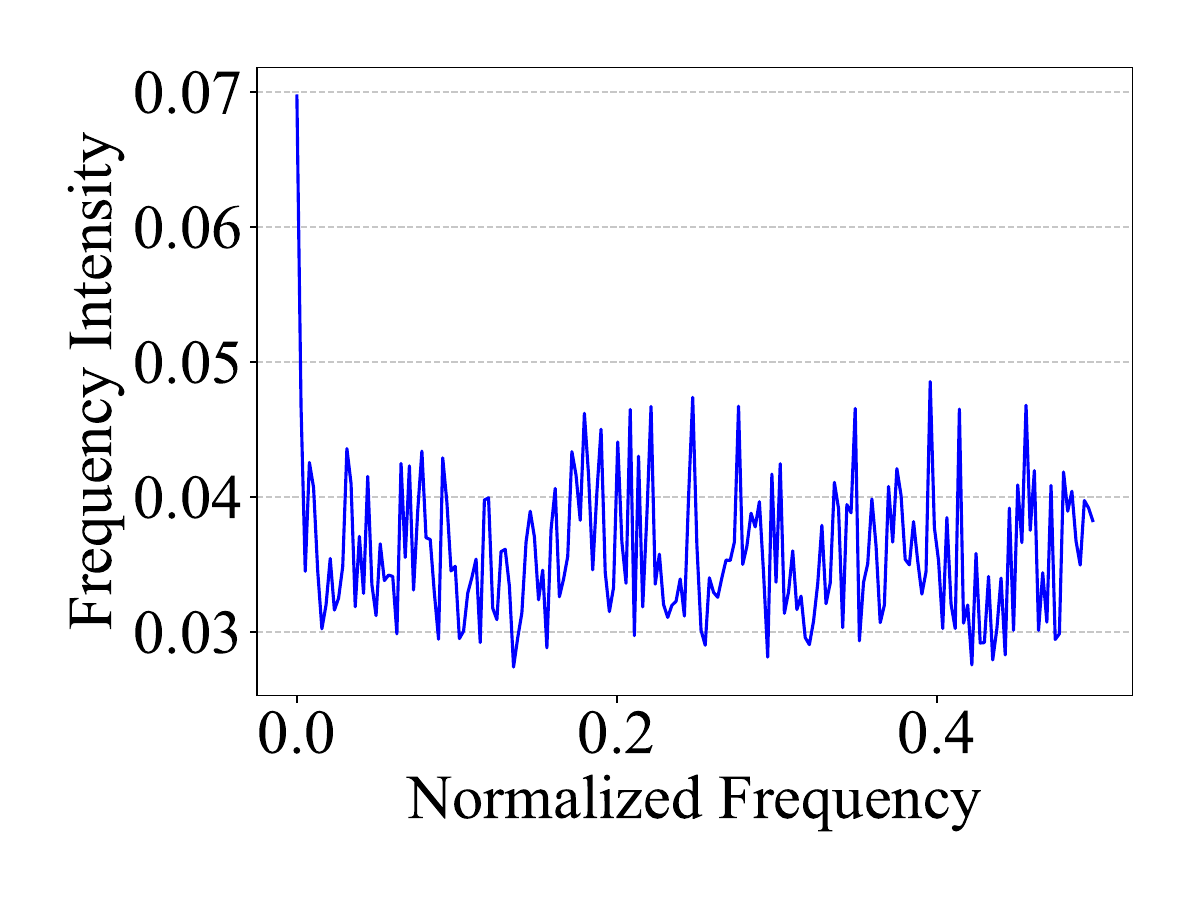}
		}
	\end{minipage}
	\begin{minipage}[b]{\linewidth}
		\subfigure[Layer $5$.]{
			\includegraphics[width=0.46\linewidth]{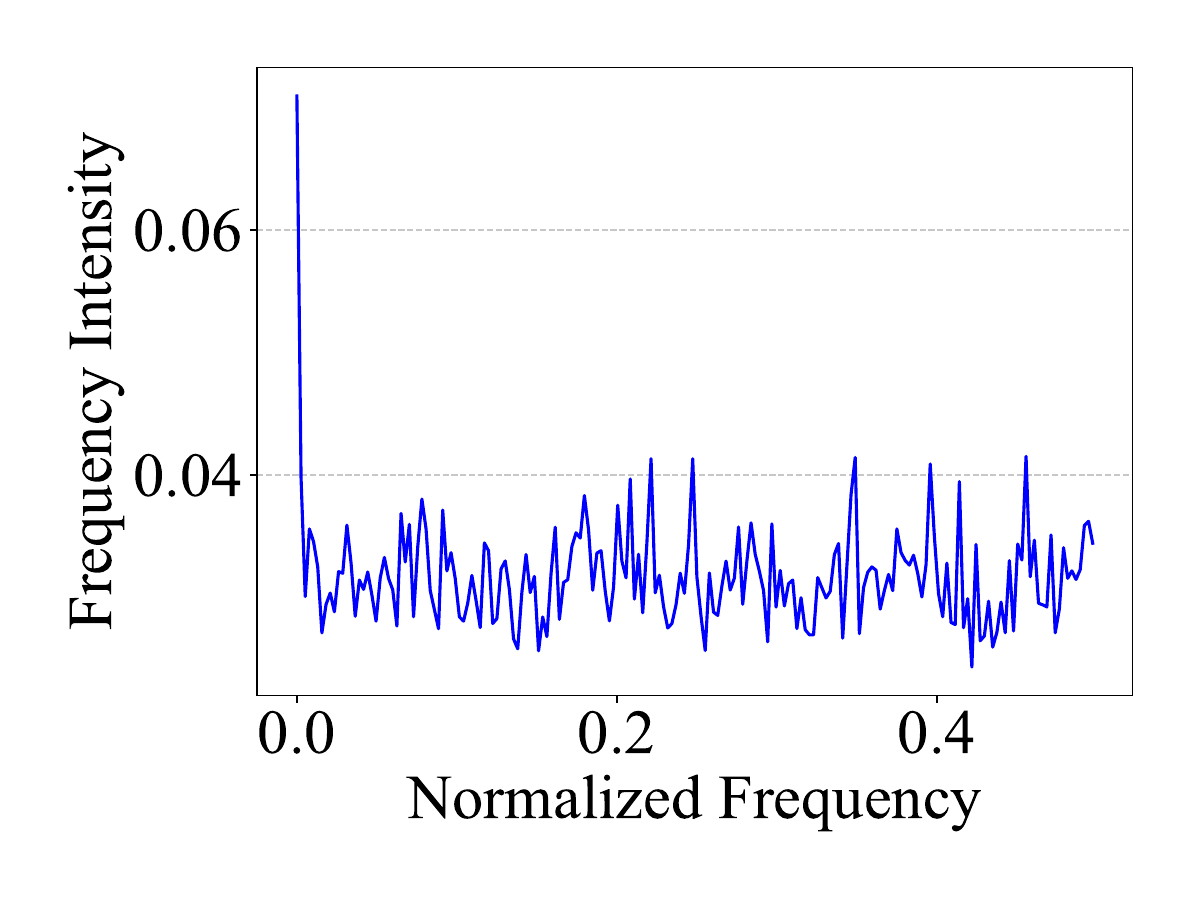}
		}
		\subfigure[Layer $6$.]{
			\includegraphics[width=0.46\linewidth]{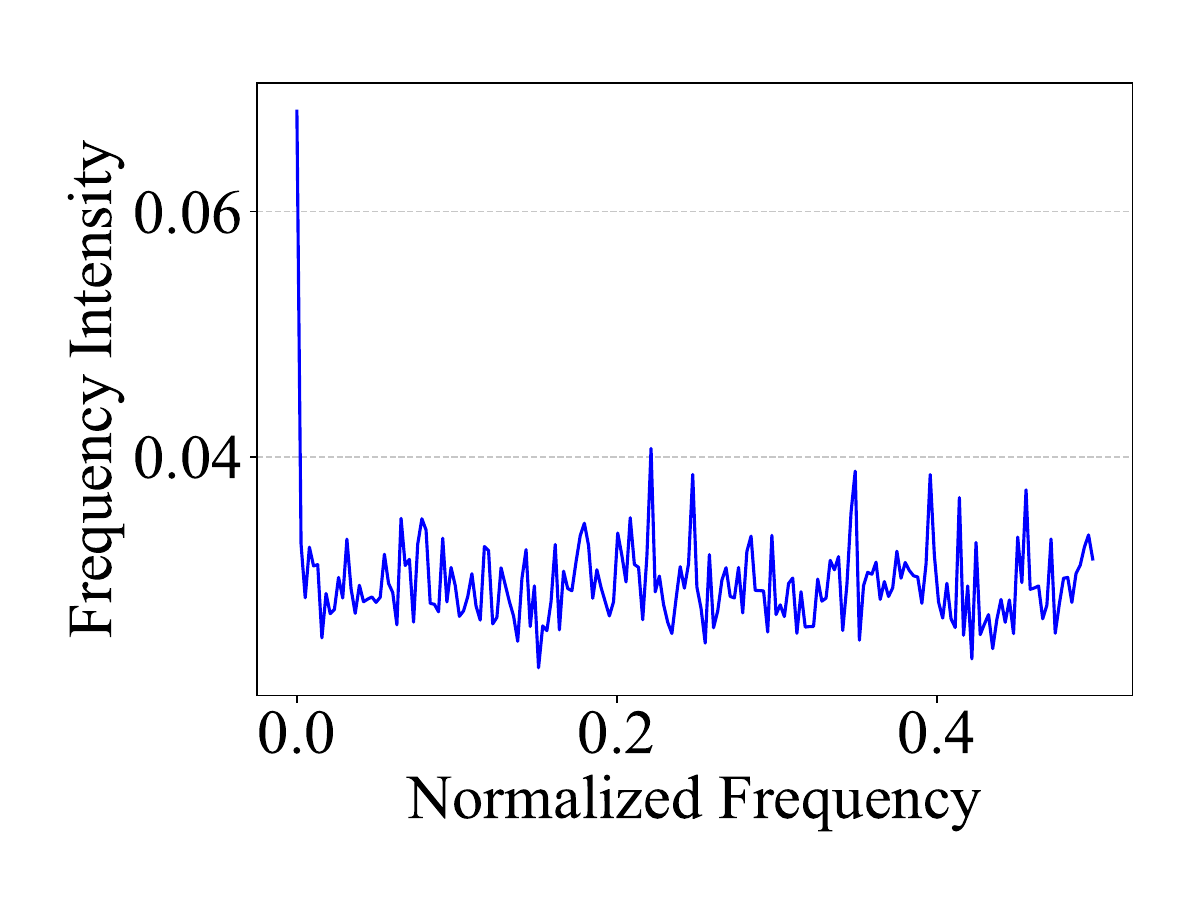}
		}
	\end{minipage}
	\caption{Spectral intensity distributions $\mathbf{S}(\mathbf{X})$ computed using Equation \eqref{channel_intensity} for layers (1-6) of CaiT-S24 feature maps.}
\end{figure}

\begin{figure}[t]
	\centering
	\begin{minipage}[b]{\linewidth}
		\subfigure[Layer $7$.]{
			\includegraphics[width=0.46\linewidth]{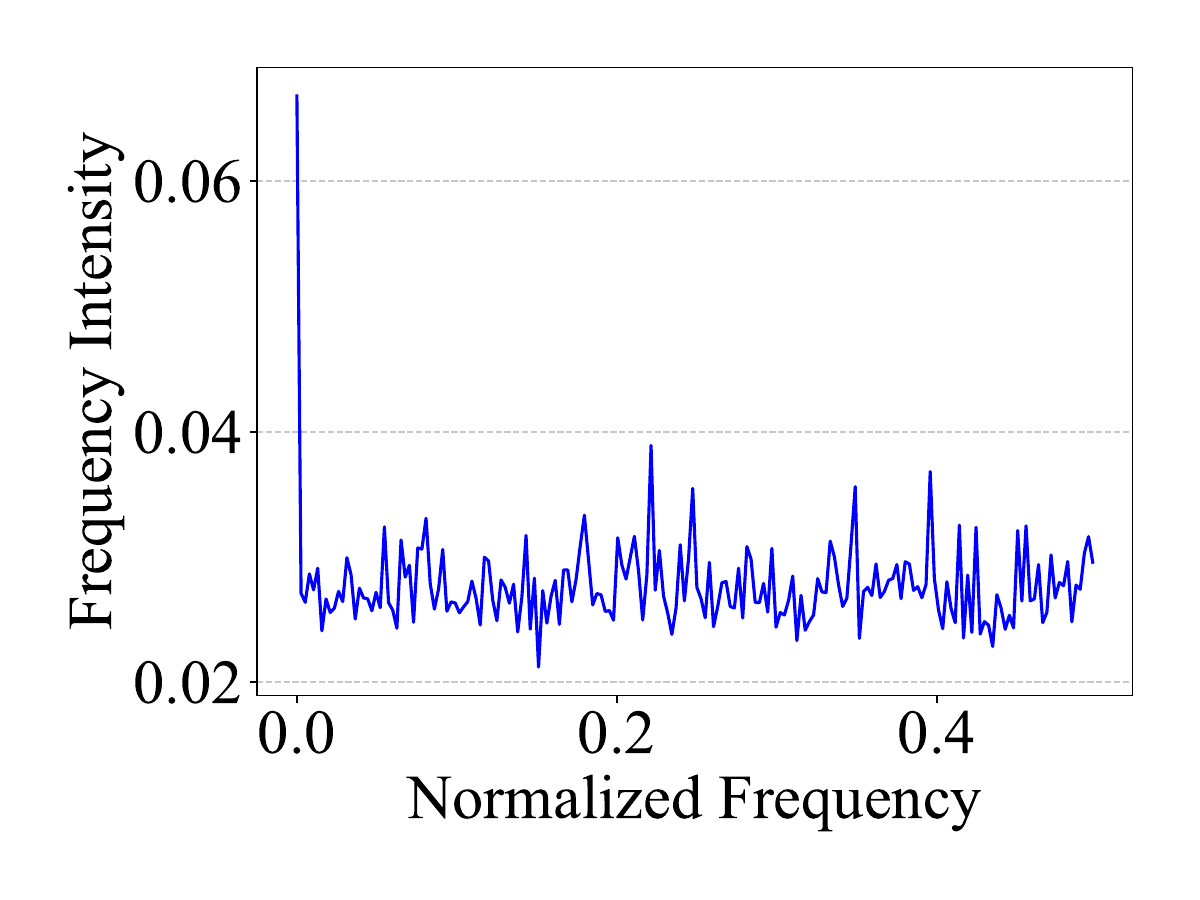}
		}
		\subfigure[Layer $8$.]{
			\includegraphics[width=0.46\linewidth]{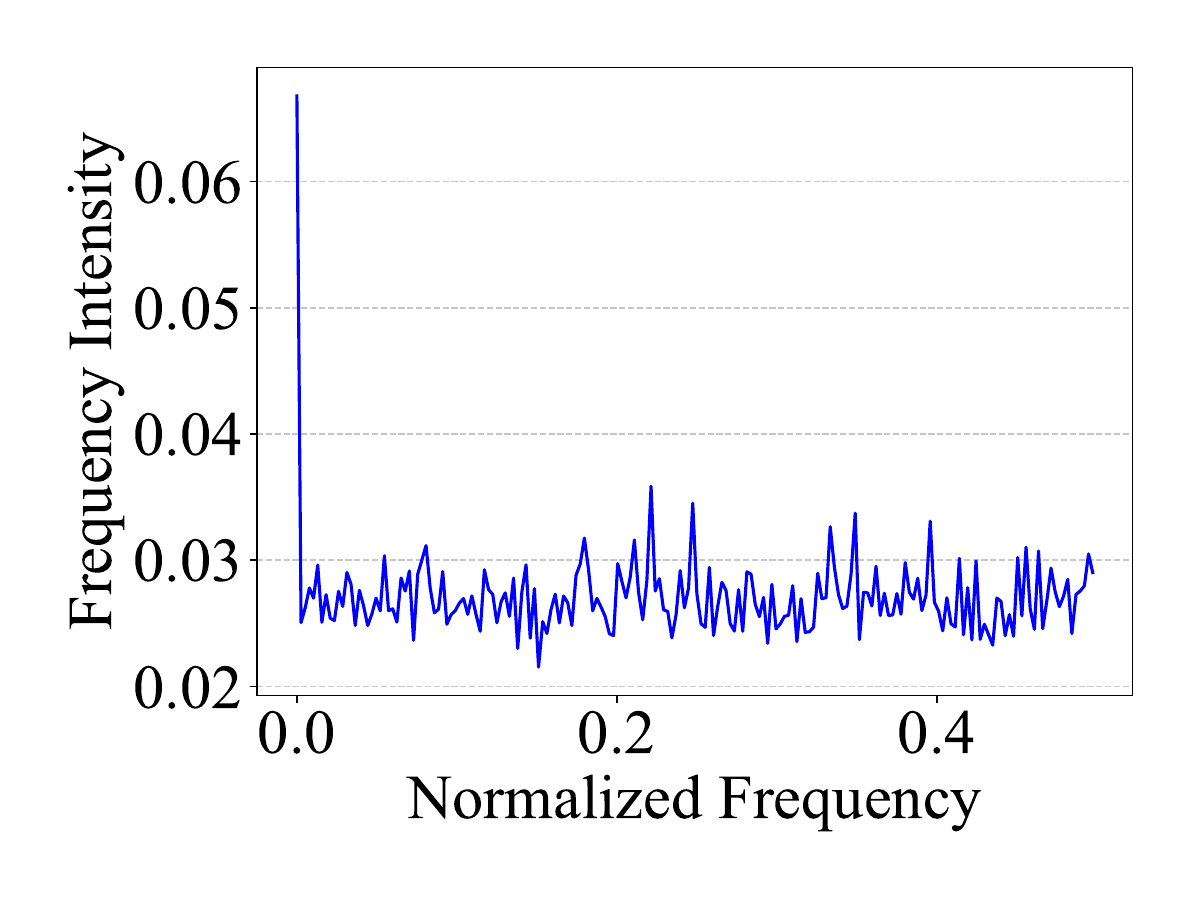}
		}
	\end{minipage}
	\begin{minipage}[b]{\linewidth}
		\subfigure[Layer $9$.]{
			\includegraphics[width=0.46\linewidth]{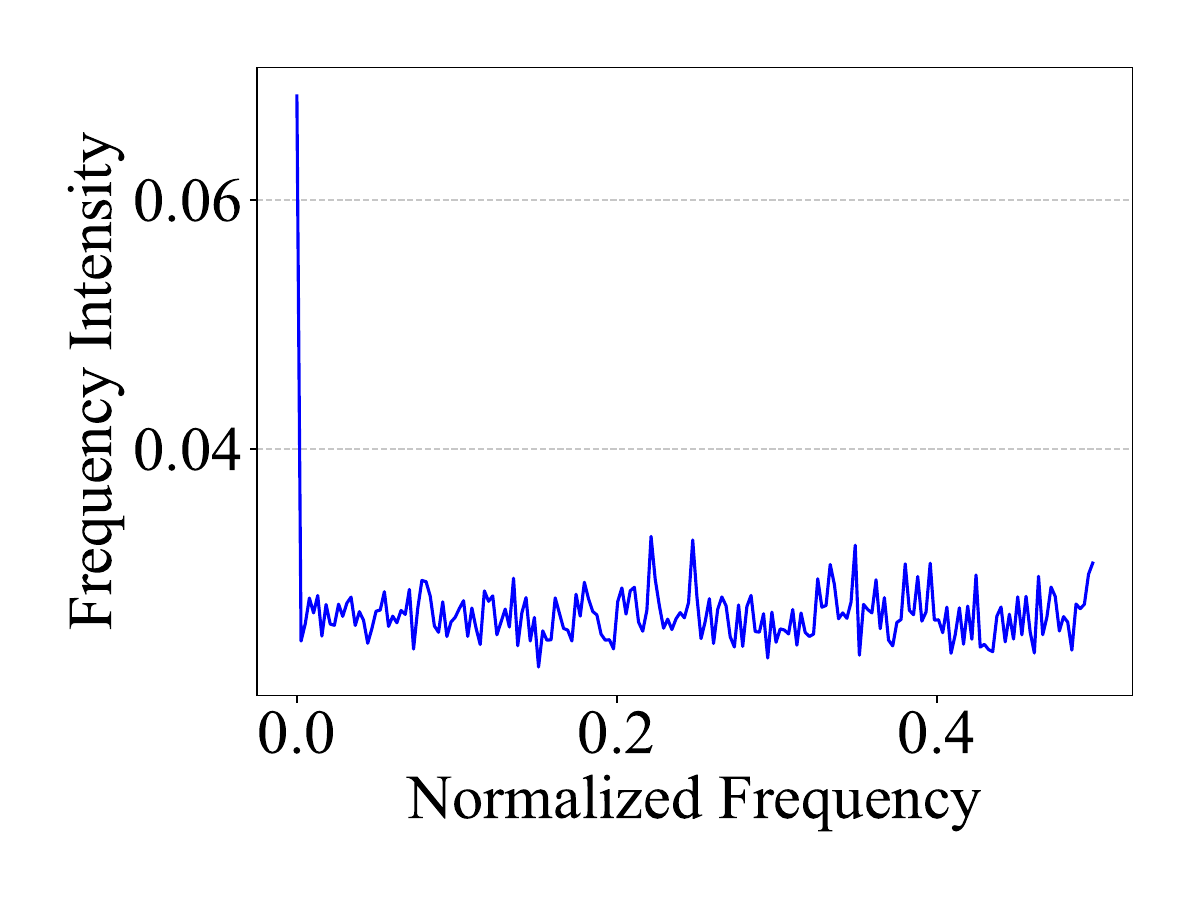}
		}
		\subfigure[Layer $10$.]{
			\includegraphics[width=0.46\linewidth]{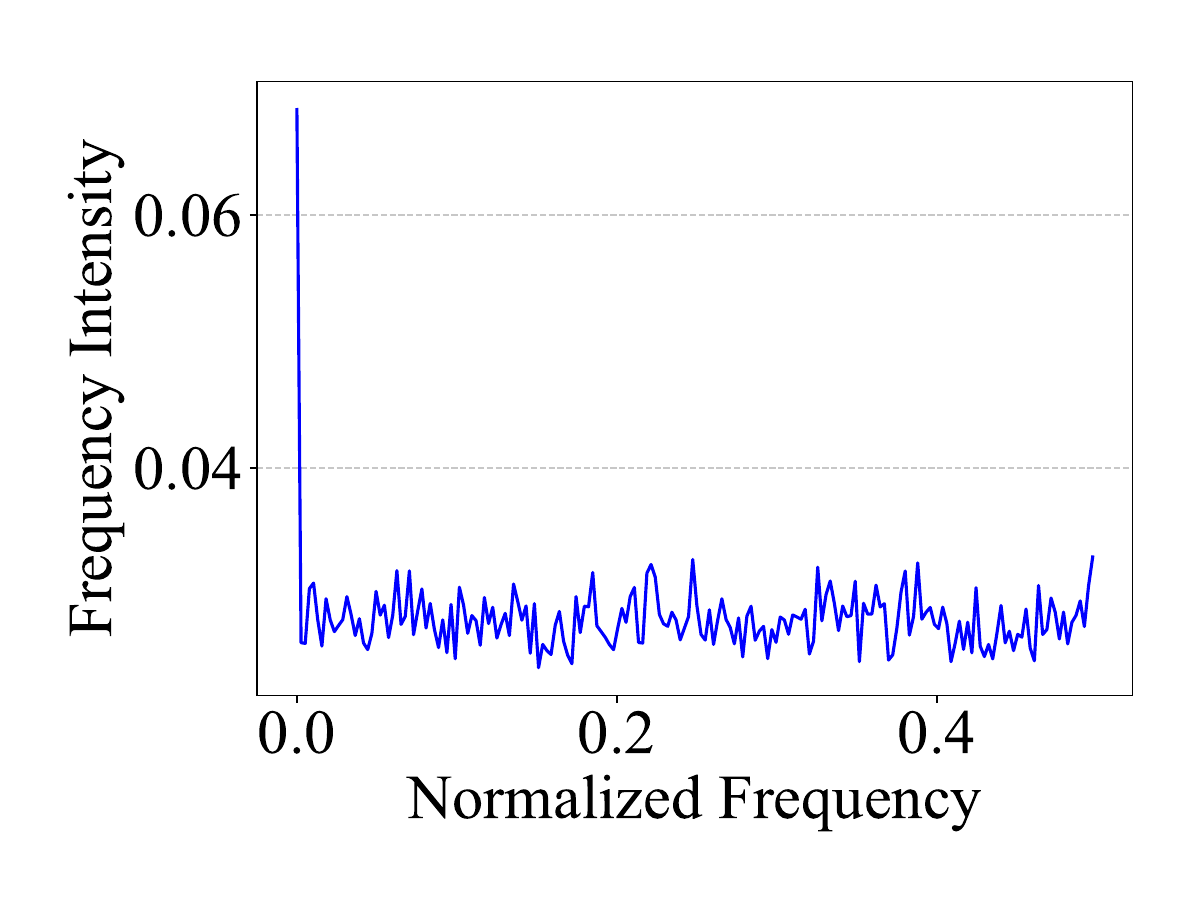}
		}
	\end{minipage}
	\begin{minipage}[b]{\linewidth}
		\subfigure[Layer $11$.]{
			\includegraphics[width=0.46\linewidth]{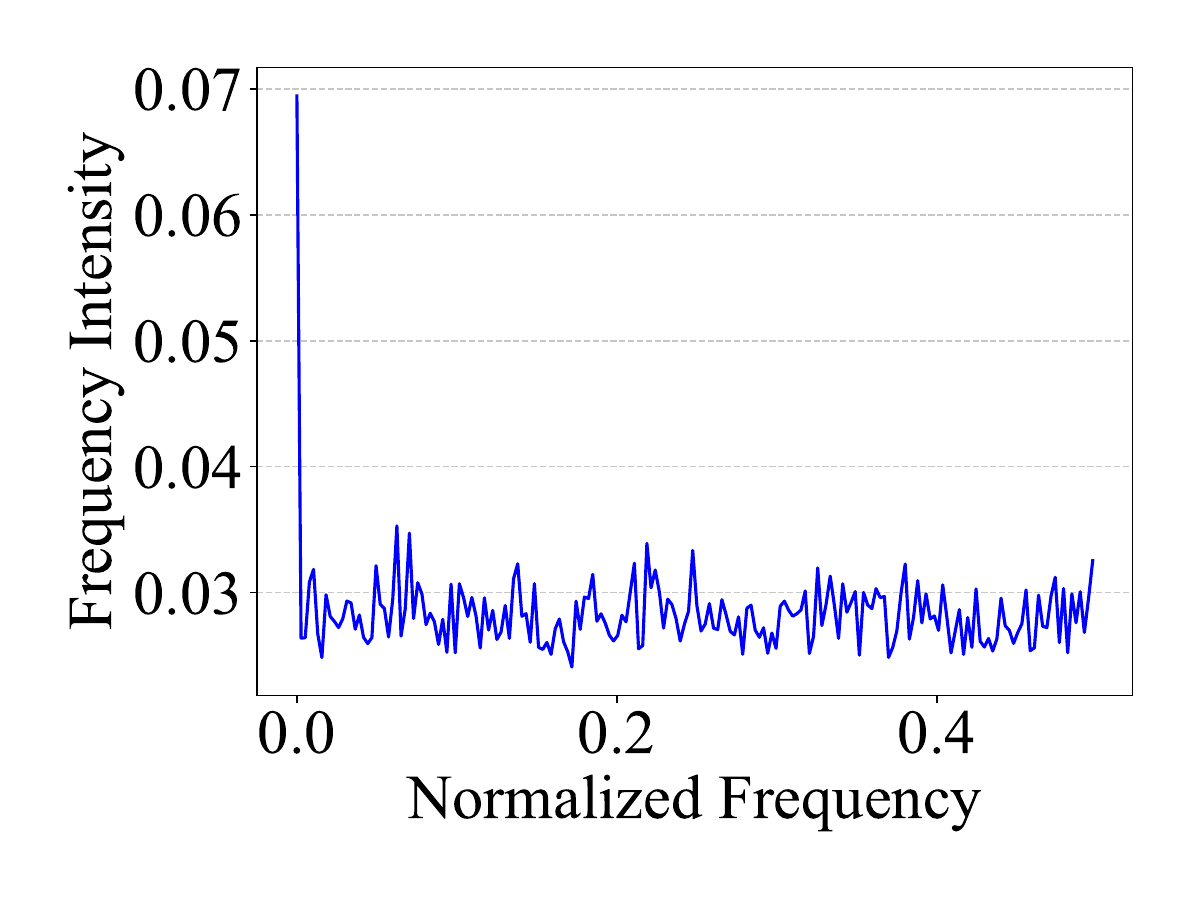}
		}
		\subfigure[Layer $12$.]{
			\includegraphics[width=0.46\linewidth]{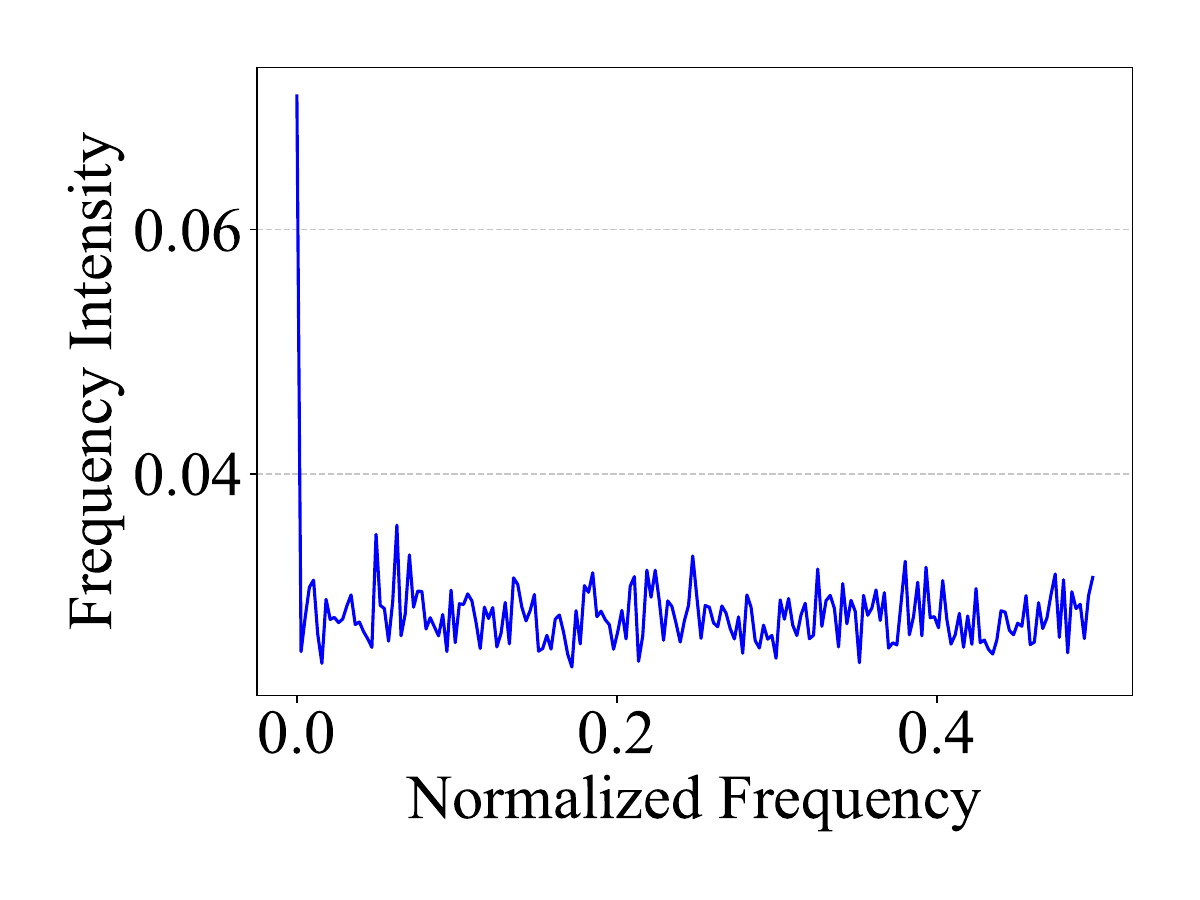}
		}
	\end{minipage}
	\caption{Spectral intensity distributions $\mathbf{S}(\mathbf{X})$ computed using Equation \eqref{channel_intensity} for layers (7-12) of CaiT-S24 feature maps.}
\end{figure}

\begin{figure}[t]
	\centering
	\begin{minipage}[b]{\linewidth}
		\subfigure[Layer $13$.]{
			\includegraphics[width=0.46\linewidth]{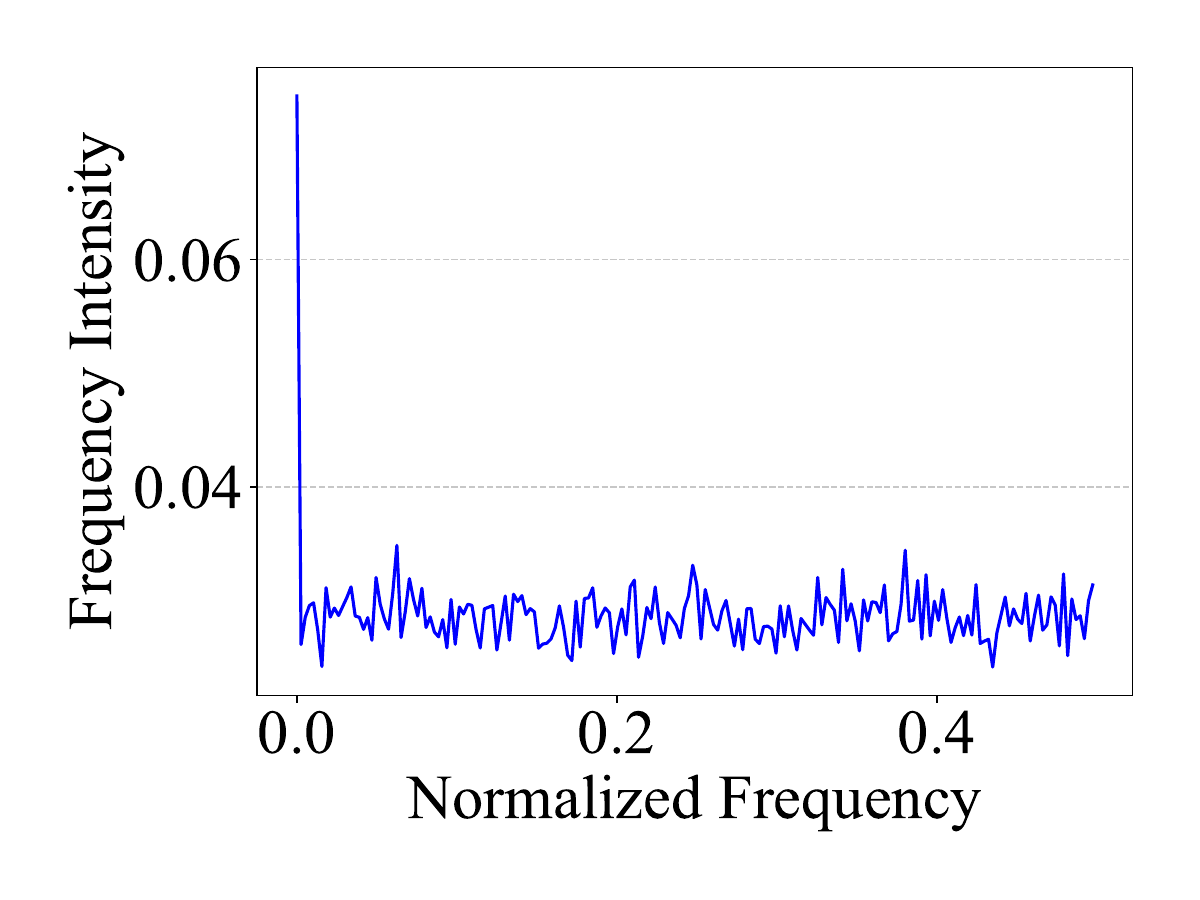}
		}
		\subfigure[Layer $14$.]{
			\includegraphics[width=0.46\linewidth]{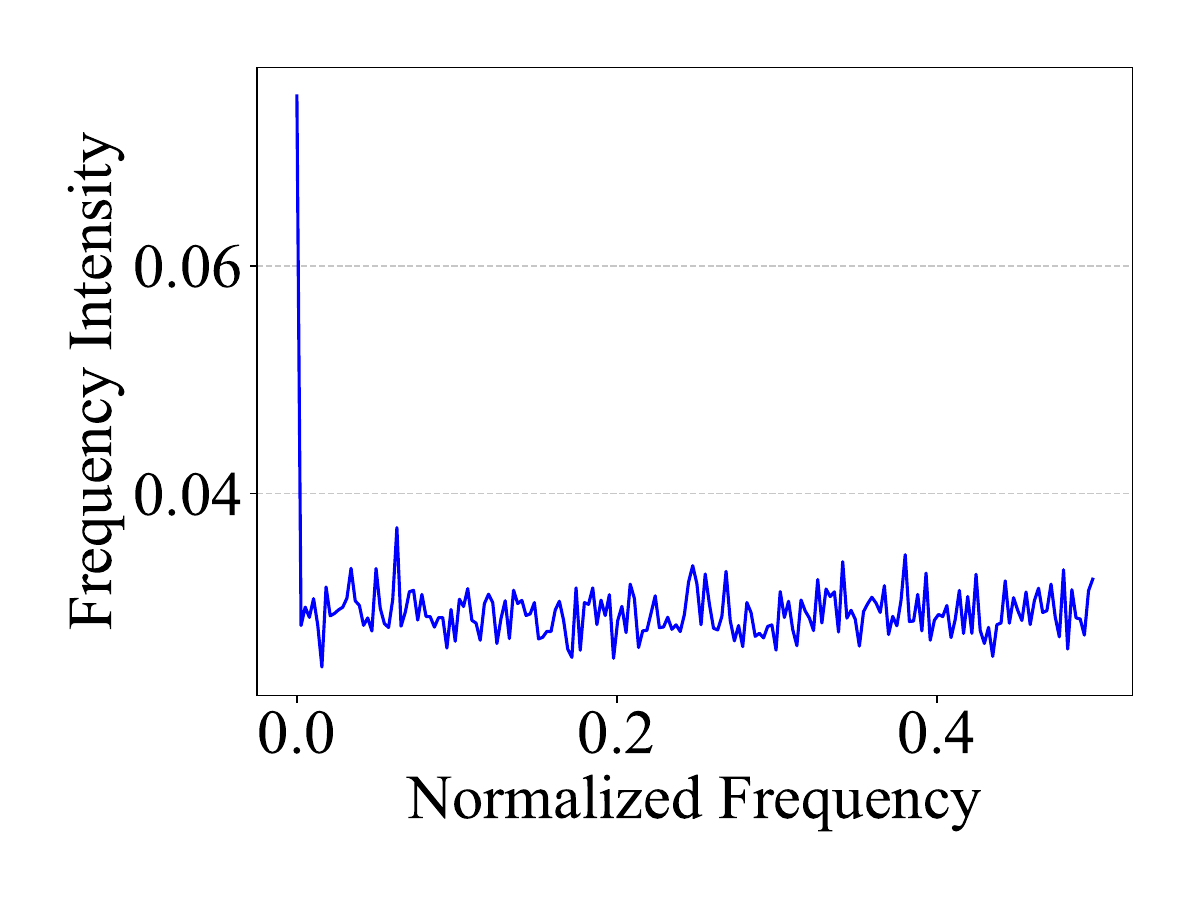}
		}
	\end{minipage}
	\begin{minipage}[b]{\linewidth}
		\subfigure[Layer $15$.]{
			\includegraphics[width=0.46\linewidth]{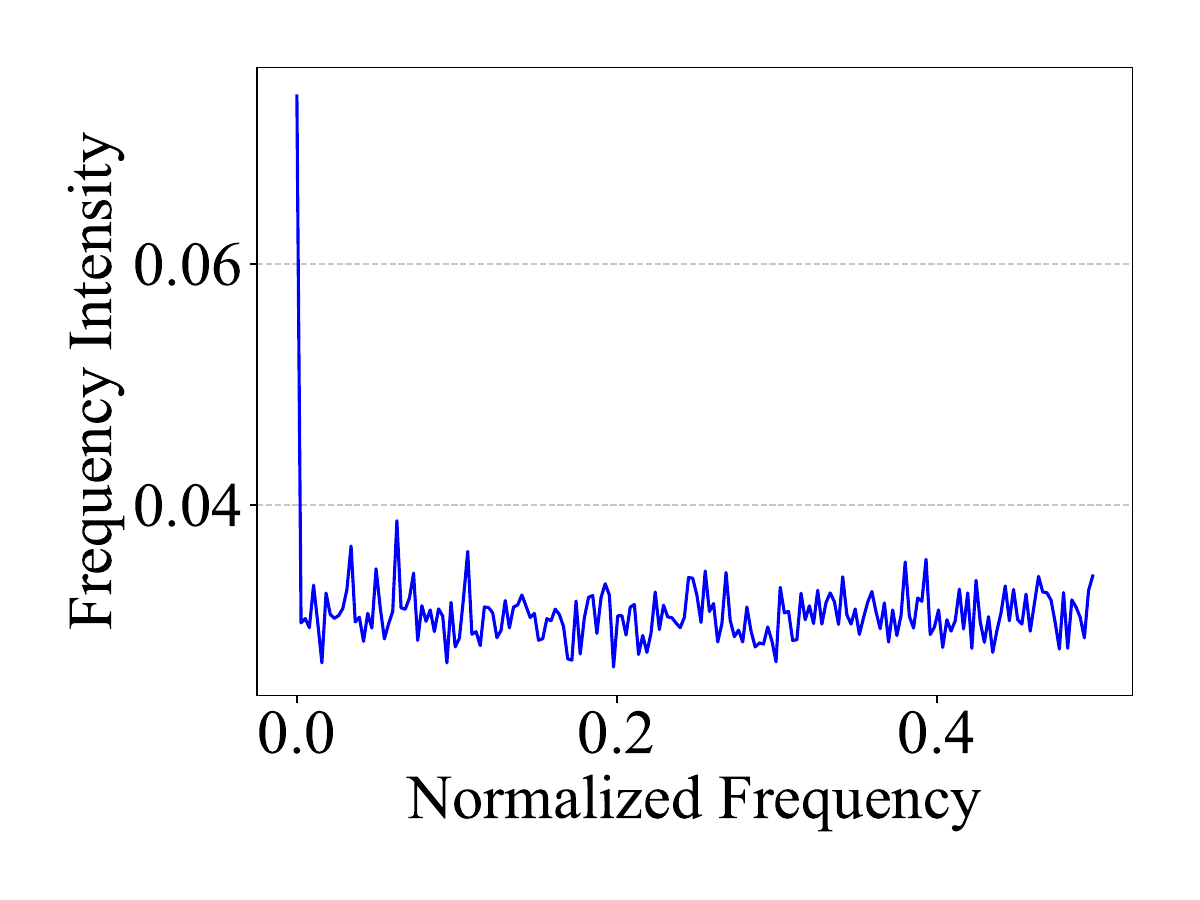}
		}
		\subfigure[Layer $16$.]{
			\includegraphics[width=0.46\linewidth]{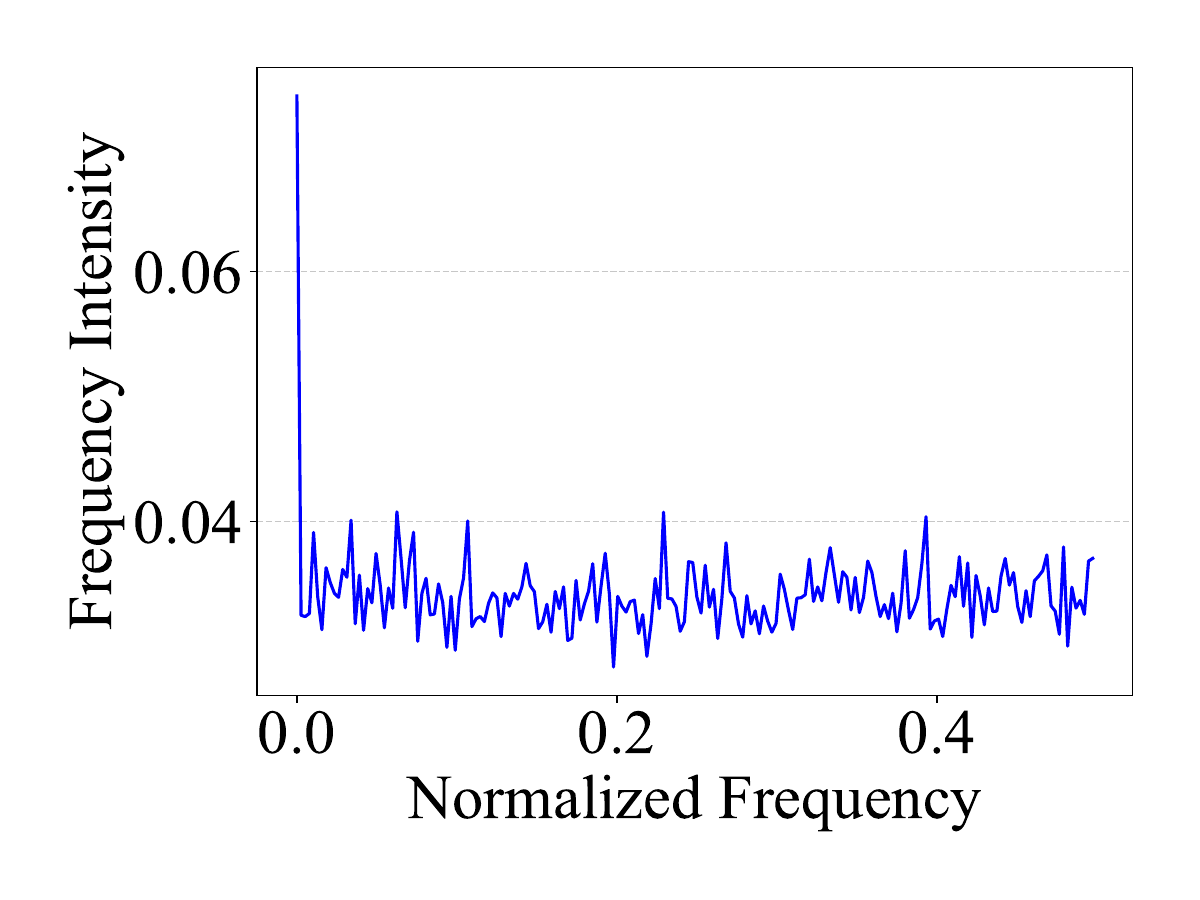}
		}
	\end{minipage}
	\begin{minipage}[b]{\linewidth}
		\subfigure[Layer $17$.]{
			\includegraphics[width=0.46\linewidth]{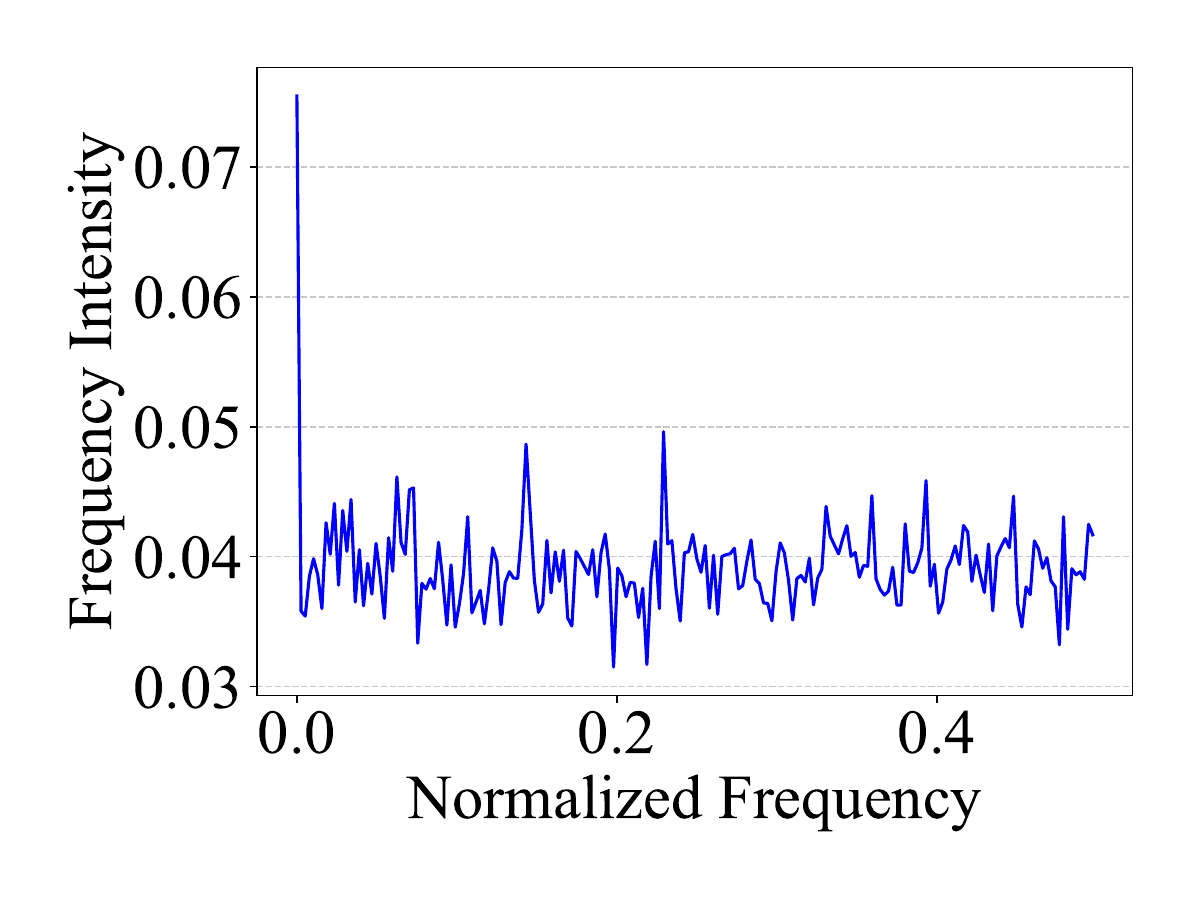}
		}
		\subfigure[Layer $18$.]{
			\includegraphics[width=0.46\linewidth]{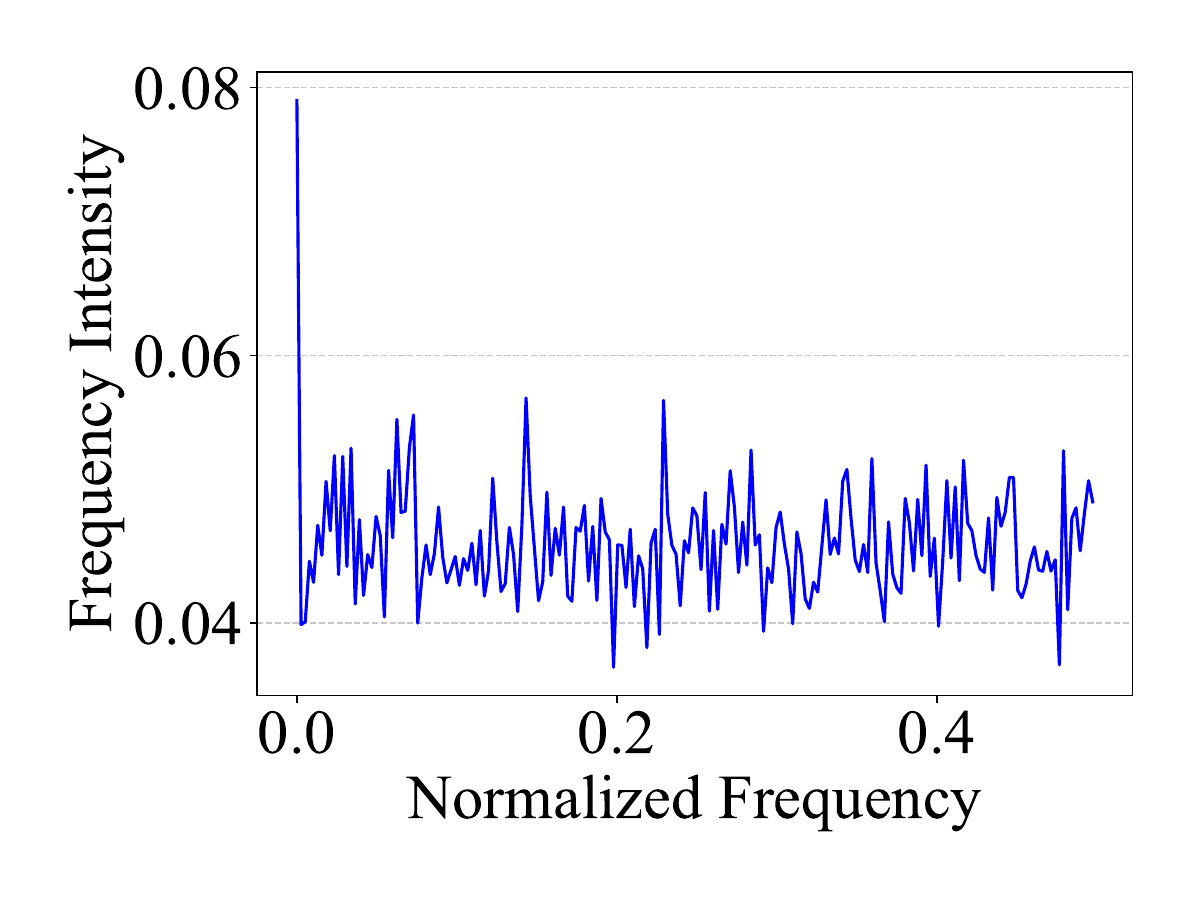}
		}
	\end{minipage}
	\caption{Spectral intensity distributions $\mathbf{S}(\mathbf{X})$ computed using Equation \eqref{channel_intensity} for layers (13-18) of CaiT-S24 feature maps.}
\end{figure}

\begin{figure}[t]
	\centering
	\begin{minipage}[b]{\linewidth}
		\subfigure[Layer $19$.]{
			\includegraphics[width=0.46\linewidth]{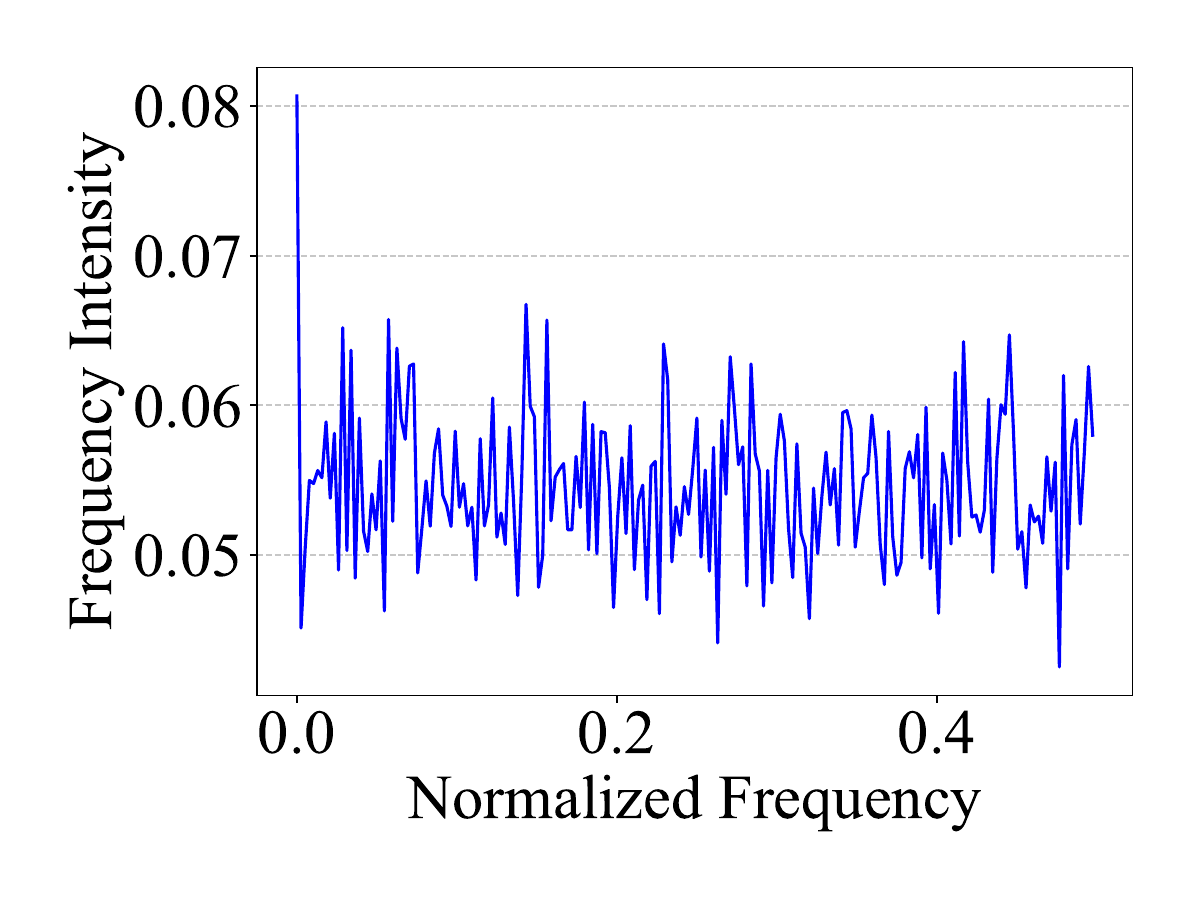}
		}
		\subfigure[Layer $20$.]{
			\includegraphics[width=0.46\linewidth]{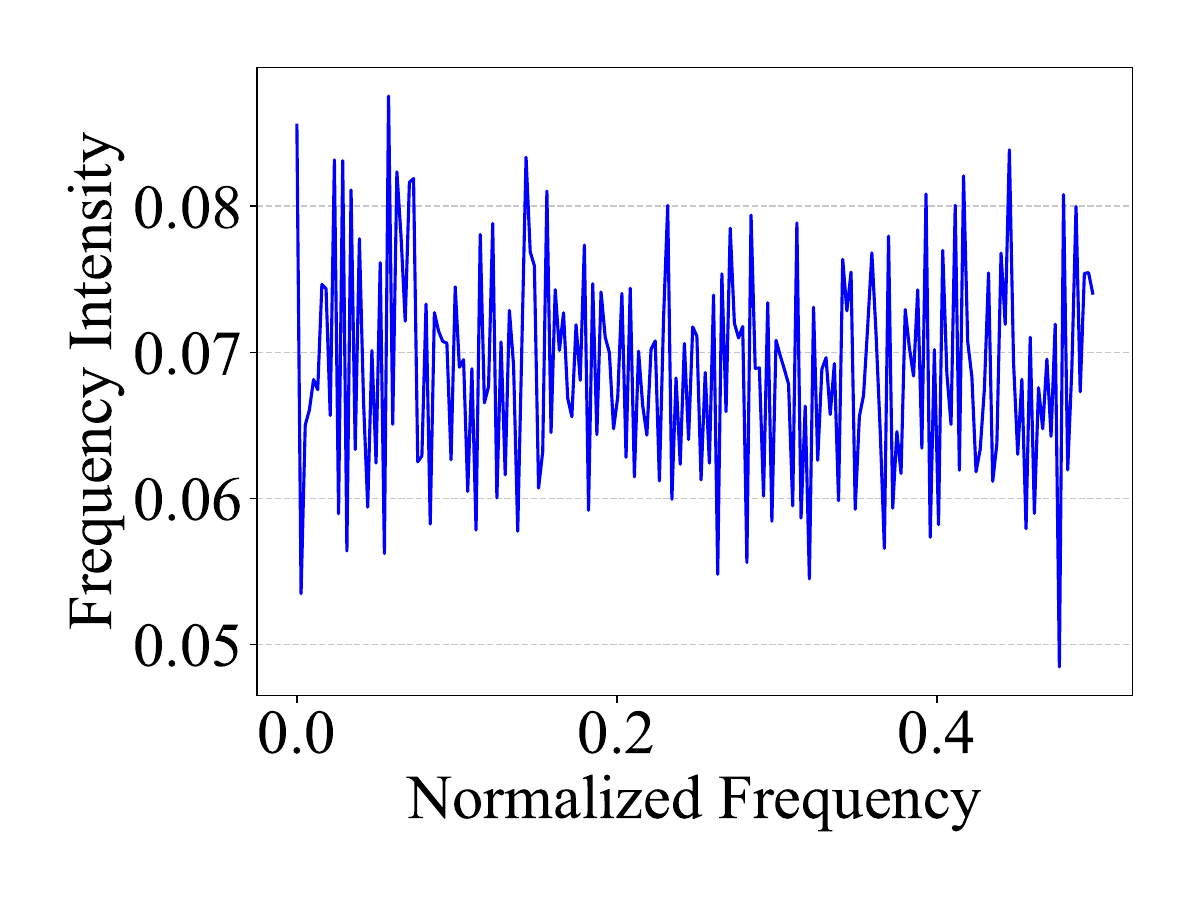}
		}
	\end{minipage}
	\begin{minipage}[b]{\linewidth}
		\subfigure[Layer $21$.]{
			\includegraphics[width=0.46\linewidth]{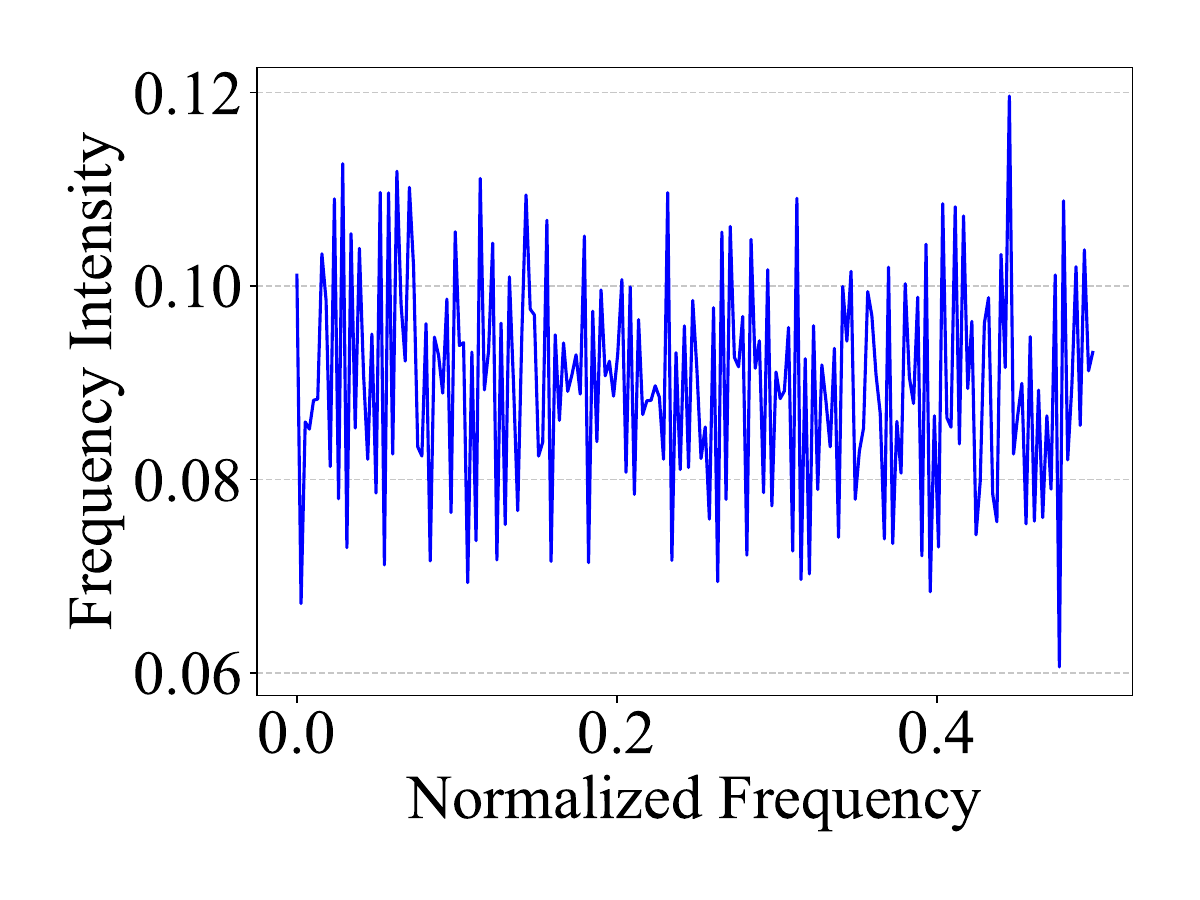}
		}
		\subfigure[Layer $22$.]{
			\includegraphics[width=0.46\linewidth]{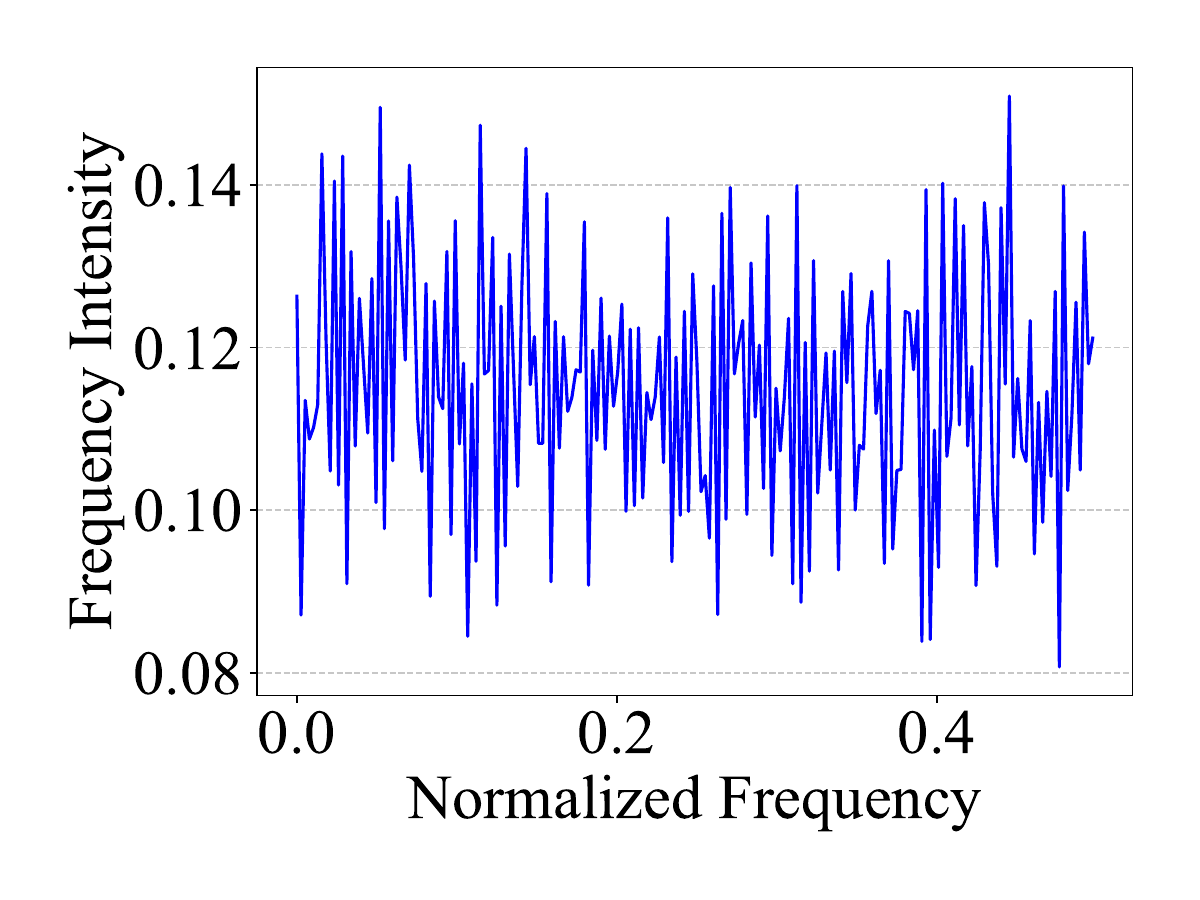}
		}
	\end{minipage}
	\begin{minipage}[b]{\linewidth}
		\subfigure[Layer $23$.]{
			\includegraphics[width=0.46\linewidth]{Frequency_Intensity_22_th_layer.pdf}
		}
		\subfigure[Layer $24$.]{
			\includegraphics[width=0.46\linewidth]{Frequency_Intensity_23_th_layer.pdf}
		}
	\end{minipage}
	\caption{Spectral intensity distributions $\mathbf{S}(\mathbf{X})$ computed using Equation \eqref{channel_intensity} for layers (19-24) of CaiT-S24 feature maps.}
\end{figure}

\end{document}